\documentclass[a4paper]{jpconf}
\usepackage{graphicx}
\usepackage{amsmath}
\usepackage{amsthm}
\usepackage{dirtytalk}
\usepackage{color}
\usepackage{mathrsfs}
\usepackage{amsfonts}
\usepackage[inline]{enumitem}
\usepackage{bbm}
\usepackage{float}
\usepackage[backend=biber,style=nature]{biblatex}
\usepackage{fancyhdr}
\usepackage{lineno}
\usepackage{hyperref}

\usepackage{mathptmx} % assumes new font selection scheme installed
\usepackage{times} % assumes new font selection scheme installed
\usepackage{amsmath} % assumes amsmath package installed
\usepackage{amssymb}  % assumes amsmath package installed
\usepackage{dirtytalk}
\usepackage{chemformula}
\usepackage{hhline}
\usepackage{makecell}
\usepackage{gensymb}
\usepackage{times}
\usepackage{fixltx2e}

\usepackage{algpseudocode}% http://ctan.org/pkg/algorithmicx
\usepackage{algorithm}% http://ctan.org/pkg/algorithm
\algnewcommand{\algorithmicgoto}{\textbf{go to}}%
\algnewcommand{\Goto}[1]{\algorithmicgoto~\ref{#1}}%

\usepackage{amsmath}

\begin{document}

\title{Robust Anthropomorphic Robotic Manipulation through Biomimetic Distributed Compliance}

\thispagestyle{fancy}% The IEEE template will automatically declare/thispagestyle{plain} after/maketitle, 
                    % Results in nothing on the first page. So we have to change plain to fancy
\lhead{}% Left of the header, if you need something, add it in {}
\chead{}% in the header
\rhead{}% header right
\lfoot{}% header left
\cfoot{\thepage}% in the header
\rfoot{}
\renewcommand{\headrulewidth}{0pt}% change to 0pt to remove the horizontal line under the header
\renewcommand{\footrulewidth}{0pt}% change to 0pt to remove the horizontal line above the footer

\pagestyle{fancy}
\cfoot{\thepage}

\author{Kai Junge,$^{1\ast}$ Josie Hughes$^{1}$}

\address{$^{1}$CREATE Lab, EPFL, Lausanne, Switzerland \\
          $\ast$ Corresponding author. Email: kai.junge@epfl.ch}

\begin{abstract}
% High level motivation
The impressive capabilities of humans to robustly perform manipulation relies on compliant interactions, enabled through the structure and materials spatially distributed in our hands.
% Hypothesis
We propose by mimicking this distributed compliance in an anthropomorphic robotic hand, the open-loop manipulation robustness increases and observe the emergence of human-like behaviours. 
% Method - robot hand
To achieve this, we introduce the ADAPT Hand equipped with tunable compliance throughout the skin, fingers, and the wrist.
% Full hand
Through extensive automated pick-and-place tests, we show the grasping robustness closely mirrors an estimated geometric theoretical limit, while `stress-testing' the robot hand to perform 800+ grasps.
% Self organization
Finally, 24 items with largely varying geometries are grasped in a constrained environment with a success rate of 93\%. 
We demonstrate the hand-object self-organization behavior underlines this extreme robustness, where the hand automatically exhibits different grasp types depending on object geometries.
Furthermore, the robot grasp type mimics a natural human grasp with a direct similarity of 68\%.
\end{abstract}

% \linenumbers

%%%%%%%%%%%%%%%%%%%%%%%%%%%%%%%%%%%%%%%%%%%%%%%%%%%%%%%%%%%%%%%%%%%%%%%%%%%%%%%%%%%%%%%%%%%%%%%
%%%  SECTION: Introduction
%%%%%%%%%%%%%%%%%%%%%%%%%%%%%%%%%%%%%%%%%%%%%%%%%%%%%%%%%%%%%%%%%%%%%%%%%%%%%%%%%%%%%%%%%%%%%%%
\section{Introduction}
% Paragraph 1
The human ability to robustly pick, place, and manipulate objects in uncertain environments is common place, and we perform it with ease~\cite{rosenbaum2012cognition}.  
Whilst our planning and sensory-motor capabilities play a key role, these compliant and robust interactions are also heavily influenced by our hand morphology, kinematics, and dynamics~\cite{nanayakkara2017role,murphy2018structure}.
From softness in the skin creating stable contact, to compliant muscle synergies\cite{santello2016hand} forming diverse ranges of grasp types\cite{feix2015grasp}, this physical intelligence can enable emergent and self-organized behaviours which result in a robust response to uncertainties in the environment\cite{seminara2023hierarchical}.
This serves as inspiration for advancing the baseline open-loop capabilities of anthropomorphic robotic manipulators. 
Such physical intelligence will enable robotic hands with fluidic, human like dynamic behaviours and is essential for fully leveraging the potential of rapidly advancing control and learning strategies~\cite{piazza2019century, pfeifer2006designing}.

% Paragraph 2
A number of different approaches have been explored for advancing open-loop robustness and versatility of anthropomorphic robotic hands.
Incorporating compliance underpins all these methodologies, with many taking a embodied intelligence\cite{pfeifer2007self}, or physical intelligence perspective\cite{miriyev2020skills}. 
One approach is to develop fully compliant hands such as the RBO Hand I/II/III~\cite{deimel2016novel,puhlmann2022rbo, deimel2013compliant}, BCL-26 Hand\cite{zhou2019soft}, and the Shorthose Hand\cite{shorthose2022design}.
Typically pneumatically operated, they have shown robustness to object geometry in grasping tasks through their continuously compliant structure.
In particular, the RBO Hand III has shown significant robustness for dexterous in-hand manipulation of a rubix cube with purely open-loop actions \cite{bhatt2022surprisingly}.
However, fully soft hands can experience some limitations in terms of the possible force application, repeatability, and agility. 
An alternative approach is tendon actuated soft-rigid hands, which are inspired by and mimic the human musculoskeletal structure and actuation mechanisms\cite{weghe2004act}.
These can leverage rolling contact condyloid and synovial joints  held together by compliant ligaments ~\cite{frost1994perspectives,toshimitsu2023getting,wadsworth1983clinical} providing natural, robust behaviour with resistance to impact and joint friction\cite{kim2019fluid,cculha2016enhancement,bern2022simulation,xu2016design}.
% Compliance/robustness in actuation
Compliance embedded within the actuation mechanism is also important for generating open-loop robustness.
Although not in an anthropomorphic form factor, a compliant differential and underactuated tendon routing can enable multi-finger grippers which can robustly grasp objects with varying geometry with a single actuator\cite{odhner2014compliant, manti2015bioinspired, dollar2006robust}. 
The theoretical framework for adaptive synergies builds upon this to similarly exploit compliance to achieve diverse behaviour for an anthropomorphic hand\cite{grioli2012adaptive, della2018toward}.
% Wrist
The requirements for compliance extends beyond the hand, with the wrist motion contributing to coordinating the behaviour of passive, i.e. non-actuated hands~\cite{gilday2023sensing}. This combination of wrist motion and bio-inspired passivity in the structure has been demonstrated to enable tasks such as dynamic and varied piano playing~\cite{hughes2018anthropomorphic} and grasping~\cite{gilday2021wrist}.

In all these varied approaches it is evident that compliance advances open-loop robustness. However, in comparison to human hands, the incorporation of compliance remains limited to distinct spatial regions.  
Furthermore, we lack metrics or approaches for the systematic assessment of the inclusion of compliance \cite{gilday2022intelligent}. 
While there are benchmarks for hands, these are often a high level task based metric which is heavily affected by the control strategy\cite{cruciani2020benchmarking,mnyusiwalla2020bin,dasari2021rb2}, object focused corresponding to a sparse fail/success metric\cite{calli2015ycb}, or focus on low level static capabilities such as grasp taxonomies\cite{feix2015grasp}, range of motion \cite{kapandji1986clinical}, and hardware features\cite{vazhapilli2019systematic}.

\begin{figure}[tb]
    \centering
    \includegraphics[width=0.95\linewidth]{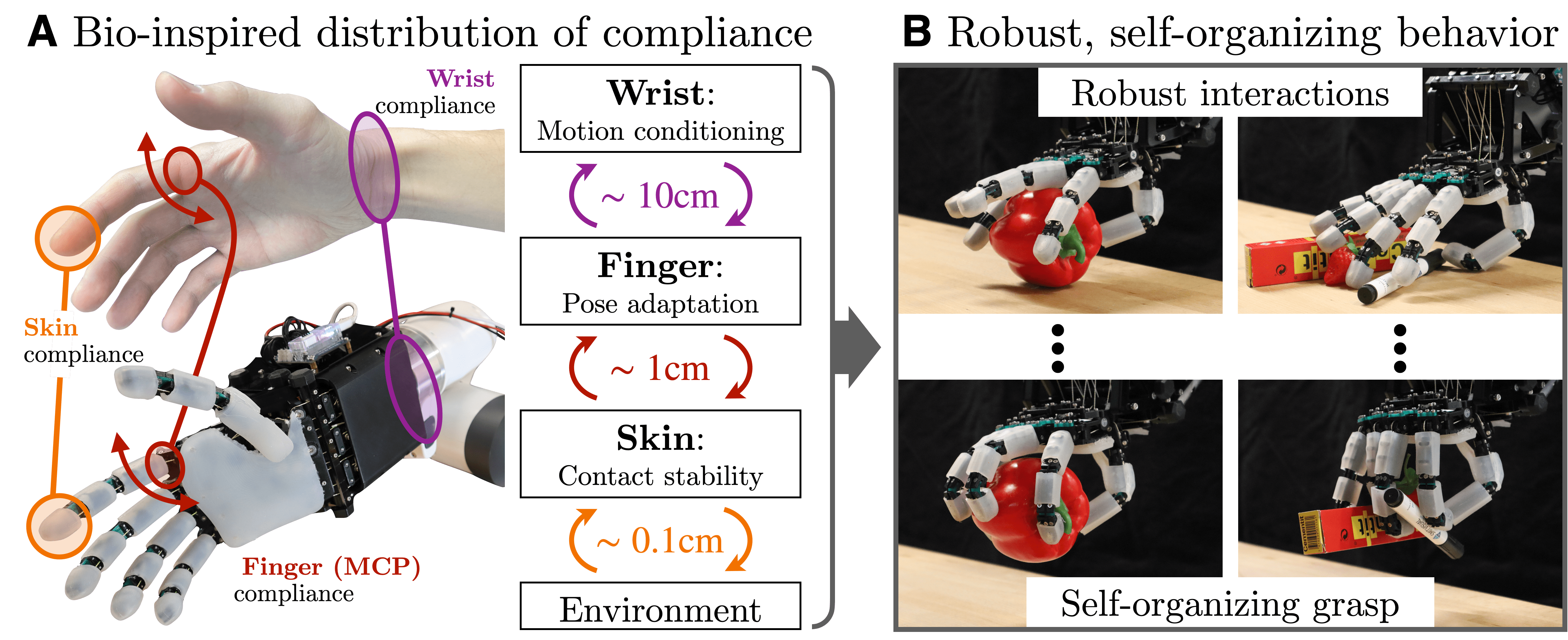}
    \caption{An anthropomorphic robot hand designed with biomimetic distribution of compliance leads to a emergence of robust and self-organizing behavior.
    A) The ADAPT Hand is compliant in its skin, finger, and wrist, which operate in the 0.1cm, 1cm, and 10cm displacement ranges.
    B) Through identical open-loop waypoints (four waypoints describing the full hand-arm motion for this grasp), the hand can grasp objects (a large apple vs three small items) successfully. The distributed compliance allows for the hand to be robust upon unknown environmental interactions, and a self-organization to take place between the hand and objects, resulting in different grasps.}
    \label{fig:fig1}
\end{figure}

We propose that matching the magnitude and distribution of compliance to that of humans is essential for matching their robustness and fluidity in interactions.  
Specifically, that by incorporating distributed and varying compliance in the skin, fingers, and wrist of an anthropomorphic robot hand, human-like behaviour can be generated with minimal open-loop control.  

A similar focus has been explored for locomotion\cite{badri2022birdbot,stella2023paws}, where it has been seen that the distribution of compliance across the body and environmental interactions leads to emergent and stable self-organizing gait patterns\cite{bourquin2004self}.
By introducing such strategically located compliance into robotic hands, we can explore the resulting hand-environment self-organization to provide robustness through \say{unconscious} selection of robust grasp configurations (see Fig.\ref{fig:fig1}B).  

This fundamentally requires a robotic hand where the compliance can be spatially distributed.  
We introduce our ADAPT (Adaptive Dexterous Anthropomorphic Programmable sTiffness) Hand, shown in Fig.\ref{fig:fig1}, which features a soft skin covering the contact surface, series elastic actuated fingers/thumb, and an impedance controlled wrist, all of which are tuned to match their biological counterpart. 
Each compliant element acts at a different interaction length scale, providing contact stability, pose adaptation, and motion conditioning.%, and facilitating human-like self-organisation for grasping.
Starting from the skin and the fingers, we demonstrate that matching the compliance to that of human hands leads to stable and robust manipulation tasks in comparison to a non-matched rigid configuration.
This leverages the tunability of the ADAPT hand, enabling us to compare and contrast bio-inspired distributed compliance and a rigid hand. 
Extending to the full hand, and using a robotic setup to autonomously perform pick-and-place tasks, we quantify the open-loop robustness for grasping approaches with respect to an estimated theoretical limit based on object and hand geometry.
Using the same setup, we also evaluate the robustness of the ADAPT Hand design through an extensive autonomous experiment with over 500 grasps across 9+ hours of uninterrupted pick-and-place operation.  The hand showed a success rate of 97\%. 
Finally, by introducing a compliant wrist we show how the distributed compliance leads to self-organization of emergent grasp types, which matches that of a human (direct similarity of 68\%), with a success rate of 93\% to handle 24 items spanning a few to 100s of millimeters.

%%%%%%%%%%%%%%%%%%%%%%%%%%%%%%%%%%%%%%%%%%%%%%%%%%%%%%%%%%%%%%%%%%%%%%%%%%%%%%%%%%%%%%%%%%%%%%%
%%%  SECTION: Results
%%%%%%%%%%%%%%%%%%%%%%%%%%%%%%%%%%%%%%%%%%%%%%%%%%%%%%%%%%%%%%%%%%%%%%%%%%%%%%%%%%%%%%%%%%%%%%%
\section{Results}
% Motivate why we want bioinspired hand with compliance, and make it clear what the result is

% Distributing compliance through the skin, wrist and fingers enables adaptation to environmental interactions at different length scales, providing robustness from the contact level which extends to grasping stability. 

\subsection{ADAPT Hand}
\begin{figure}[tb]
    \centering
    \includegraphics[width=0.8\linewidth]{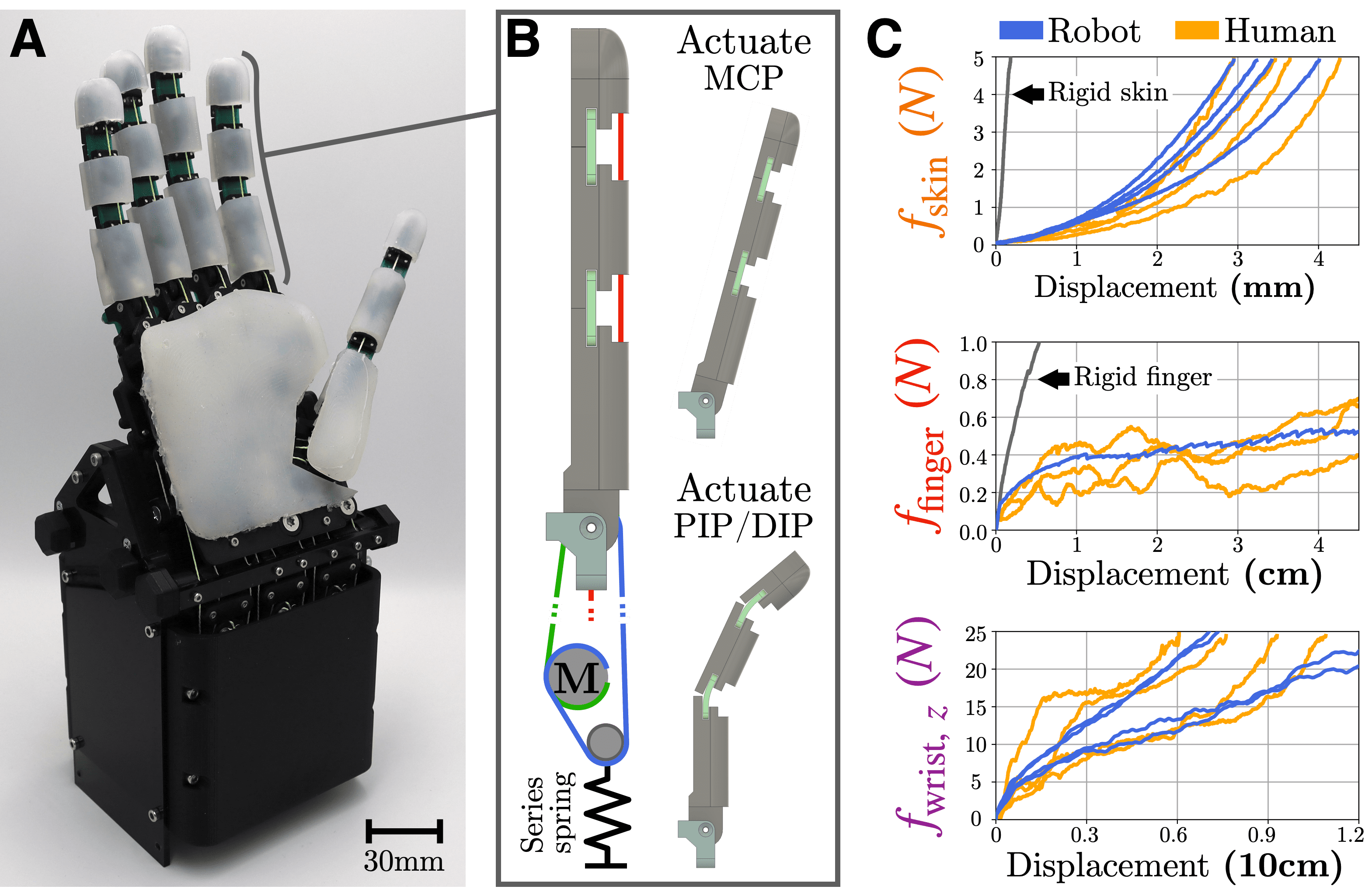}
    \caption{A) The ADAPT hand. B) Finger design with two independent actuation and a series spring on the MCP flexor. C) The force-displacement measurement curve for the skin, finger, and wrist for the robot and a human. For the skin and finger, the rigid configuration of the ADAPT Hand is also shown for comparison.}
    \label{fig:bio-inspiration}
\end{figure}

The ADAPT Hand is a bio-inspired anthropomorphic compliant robotic hand platform, with joint kinematics that reflect an human adult's hand (detailed in Section \ref{sec:meth_adapthand}) and the compliance reflecting a relaxed human adult's hand. 
The hand is designed to have independently tunable compliance in the skin, finger, and wrist used to investigate the impact of spatially distributed compliance on the robustness and performance of manipulation tasks.
The hand, shown in Fig.\ref{fig:bio-inspiration}A, has 12 actuators corresponding to 20 degrees of freedom, and leverages tendon driven operation. 
The defining feature of the robot hand design is the finger (shown in Fig.\ref{fig:bio-inspiration}B), where two actuators control the flexion-extension action.
Two antagonistic tendons actuate the metacarpophalangeal(MCP) joint, while a single tendon is responsible for the flexion motion of the proximal inter-phalangeal(PIP) and distal inter-phalangeal(DIP) joints. 
The two actuators controls the finger motion independently (see Fig.\ref{fig:bio-inspiration}B right).
The flexor tendon of the MCP joint has a series spring attached, which provides the finger compliance. 
The abduction-adduction motion of the fingers are coupled (controlled by one actuator) and series elastic. 
The thumb is designed in a similar way, but has two antagonistic joints at the carpometacarpal(CMC) joint (base of the thumb). Despite being underactuated, the actuated range of motion can achieve all 33 grasp taxonomies\cite{feix2015grasp} and all 10 positions on the Kapandji score\cite{kapandji1986clinical} (see Fig.\ref{suppfig:taxonomy}).

Fig.\ref{fig:bio-inspiration}C shows a force-displacement curve for each type of compliance for the robot and the human. The variability of different curves for the human is resultant of multiple measurement trials. 
The skin compliance is tuned through the selection of materials to cover the underlying rigid structure. 
The chosen skin (EcoFlex20) matches that of a human, especially in comparison to a fully rigid skin (10 to 40 times stiffer).
The finger stiffness is tuned by swapping the aforementioned series spring in the MCP joint. 
Compared to the human-matched finger, the average stiffness of a rigid finger (which is realised by removing the series spring) is a factor 30 larger.
The plateauing force profile (on the human and tuned finger) from a low stiffness joint means the fingertip exerts near-constant forces under large displacements - suggesting the open-loop force application will be robust to large disturbances.
The wrist compliance is achieved by an impedance controlled robot arm. The force profile of the compliance in the vertical direction is shown, where the impedance control parameter is tuned to match a human wrist.

Throughout this work, the actuation signals are manually programmed open-loop signals. That is, an operator will record key waypoints for both the finger/thumb joints and the wrist 6dof pose (detailed in Section \ref{sec:meth_teaching}). 
During execution, the recorded waypoints are linearly interpolated and directly replayed. 
For every task in the work, the motion is programmed to best mimic how a human would perform them.

%%%%%%%%%%%%%%%%%%%%%%%%%%%%%%%%%%%%%%%%%%%%%%%%%%%%%%%%%%%%%%%%%%%%%%%%%%%%%%%%%%%%%%%%%%%%%%%
%%%  Sub-Section: Skin interaction
%%%%%%%%%%%%%%%%%%%%%%%%%%%%%%%%%%%%%%%%%%%%%%%%%%%%%%%%%%%%%%%%%%%%%%%%%%%%%%%%%%%%%%%%%%%%%%%
\subsection{Skin: Contact stability}
\label{sec:skin}
\begin{figure}[tb]
    \centering
    \includegraphics[width=0.95\linewidth]{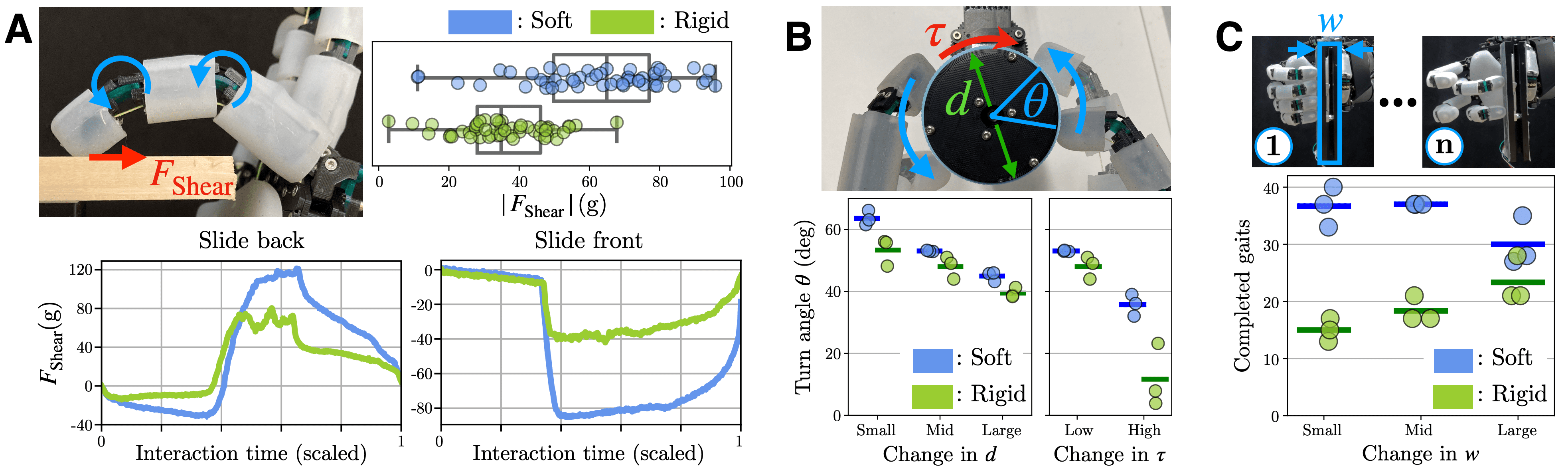}
    \caption{A) Results from the finger sliding experiment. Schematic (top-left), box plot and raw values for the shear force of rigid and soft skins measured at the midpoint of the interaction (top-right), example time series of the shear force of rigid and soft skins for the two sliding motions (bottom).
    B) Results from the knob turning experiment. Schematic (top) and turn angles for the soft and rigid skins as the diameter $d$ and resisting torque $\tau$ is varied (bottom).
    C) Results from the finger gaiting experiment. Schematic (top) and completed gaits for soft and rigid skins as the held block width $w$ is varied (bottom).
    }
    \label{fig:skin}
\end{figure}

The skin plays a crucial role in the robot-environment interaction, being the direct interface between the hand and the environment. 
The cutaneous compliance in human skin offers increased local deformation, assisting with contact stability and shear force generation.  
To measure the corresponding contact stability offered by matching the skin stiffness we contrast the matched skin to a rigid one made from PLA. 
The PLA skin is coated in a thin layer of EcoFlex20 to ensure the surface friction properties are consistent and we isolate the effect of compliance.

The main impact of the skin compliance is the increased shear forces generated for a given motion. This shear force was analyzed for two sliding motions on a plate (slide front and back motions: refer to Fig.\ref{suppfig:finger}) for different relative displacements between the finger and the sliding surface.
On the top right of Fig.\ref{fig:skin}A the shear force measured via a loadcell at the midpoint of the interaction motion is plotted. Across a range of motions and surface environments, the shear force is constantly higher with the soft skin in comparison to the rigid case. 
The bottom of Fig.\ref{fig:skin}A shows two representative shear force profiles for the two sliding motions. 
The temporal force profile is similar for the soft and rigid, but the generated force has typically twice the force magnitude.

\subsubsection{Skin compliance on task performance}
This demonstrated capacity of the soft skin to generate higher shear force under the same action contributes to contact level stability. 
To measure the resulting performance, we use the task of turning a cylindrical knob with the middle finger and thumb with a predefined finger motion.  
The turn angle $\theta$ is used as the performance metric, while the environment is varied by independently varying the diameter $d$ and resistance torque $\tau$ of the knob (Fig.\ref{fig:skin}B top). 
Resultant turn angles for each environmental variation are shown in the bottom of Fig.\ref{fig:skin}B where across all tasks the soft skin is outperforming. 
In particular, variations in $\tau$ impliy that the soft skin is more robust, shown by a a far greater performance drop for the rigid skin of 37$\deg$ compared to 18$\deg$ for the soft skin.

The effect of the softer skin also extends to tasks explicitly requiring contact stability, such as finger gating when holding an object, as the width $w$ of the block is varied from low to high. 
For a given, pre-determined finger gating pattern, the performance is measured by counting the number of completed gaits until failure. 
The results given in Figure Fig.\ref{fig:skin}C show that the soft skin leads to greater contact stability, and thus a higher performance.  
The rigid one shows increasing performance as $w$ increases, as a higher width results in higher holding forces. 
For the soft skin, there is in fact a reduction in performance with the largest width object, however, this is still almost twice the performance of the rigid setting.

In both these experiments, the soft skin creates a larger contact area with the environment which results in a more stable contact, as measured through the higher stabilizing shear forces.  This leads to consistently higher task performances and in some cases increased robustness to environmental uncertainty.

%%%%%%%%%%%%%%%%%%%%%%%%%%%%%%%%%%%%%%%%%%%%%%%%%%%%%%%%%%%%%%%%%%%%%%%%%%%%%%%%%%%%%%%%%%%%%%%
%%%  Sub-Section: Finger interaction
%%%%%%%%%%%%%%%%%%%%%%%%%%%%%%%%%%%%%%%%%%%%%%%%%%%%%%%%%%%%%%%%%%%%%%%%%%%%%%%%%%%%%%%%%%%%%%%
\subsection{Finger: Pose adaptation}
\label{sec:finger}
\begin{figure}[tb]
    \centering
    \includegraphics[width=0.99\linewidth]{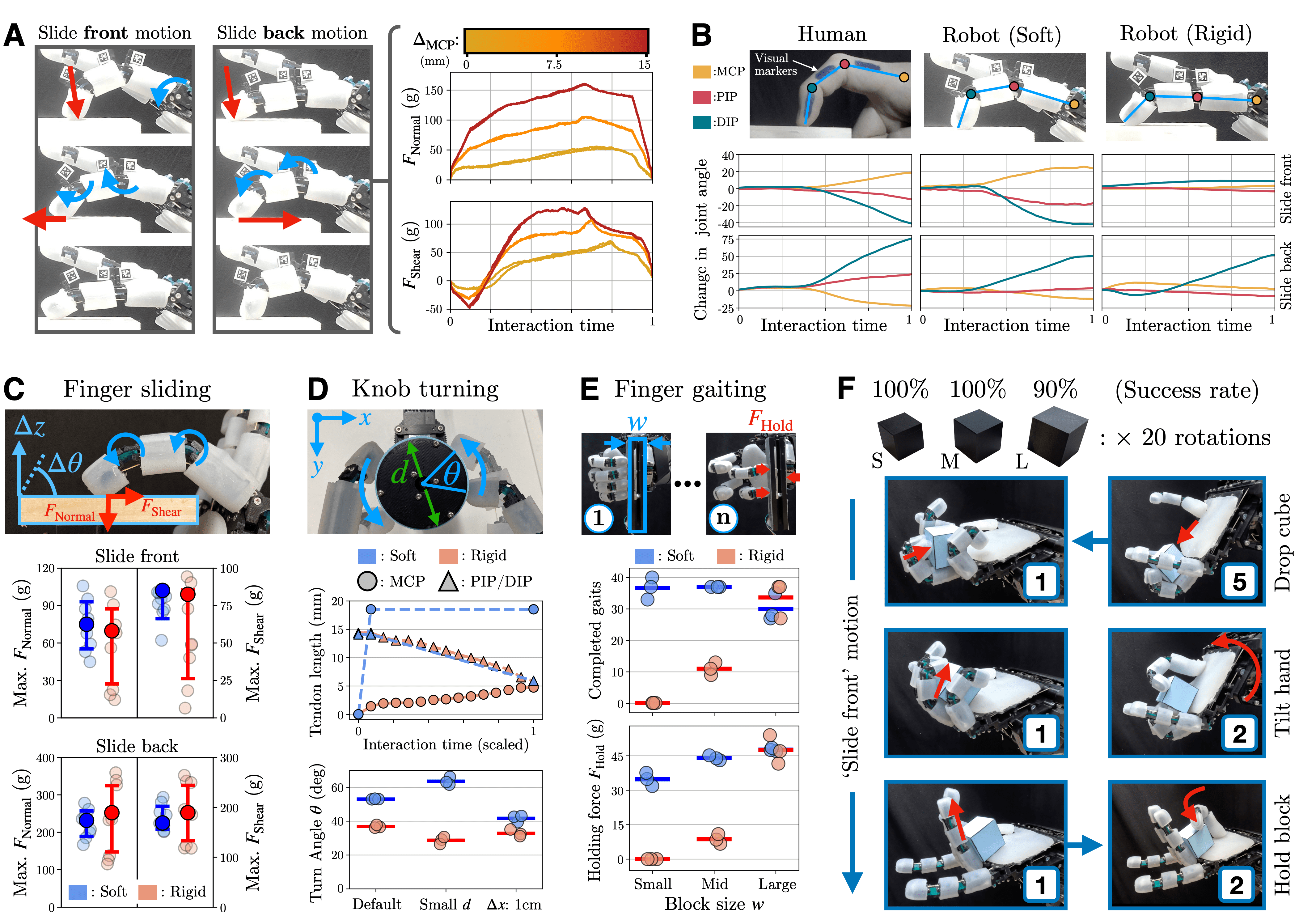}
    \caption{A) Key frames from two finger sliding motions(left). By overdriving the MCP joint, pseudo force control is possible with good repeatability (right).
    B) Trajectories of relative changes in the three joint angles for the sliding motions executed by a human, soft robot finger, and a rigid robot finger.
    C) Schematic for the finger sliding experiment (top). Maximum forces recorded as the sliding plate is displaced in the $z$ and $\theta$ directions for the two motions/soft and rigid fingers (bottom).
    D) Schematic for the knob turning experiment (top). Tendon waypoints required for the motion(mid) and turn angles as the environment is varied (bottom) for the soft and rigid fingers
    E) Schematic for the finger gaiting experiment (top). Completed gaits (mid) and average holding forces (bottom) for soft and rigid fingers as the held block width $w$ is varied.
    F) Success rates for the three cubes rotated in-hand (top). Pictorial sequence of the in-hand manipulation sequence (bottom).}
    \label{fig:finger}
\end{figure}

% 1. Conform to the surface with minimal control poses
% 2. Force control with overdrive and this is 'linear' 

The flexion-extension motion of the finger originates from the series elastically actuated MCP joint. In this section we show how the series elastic MCP joint can enable centimeter scale pose adaptation of the finger, resulting in robust behavior.

When external contact forces are applied the entire finger will adaptive comply to the environment since the MCP joint is located at the stem of the finger.
By combining this series elastic actuated joint with the PIP/DIP actuation, the fingertip can move to follow a surface with consistent contact.
For example, Fig.\ref{fig:finger}A shows two sliding motions (slide front and slide back), achieved by flexing or extending the PIP/DIP joints respectively, which is achieved through three waypoint transitions (detailed in Fig.\ref{suppfig:finger}). 
The compliant MCP mechanism also offers pseudo force-control, by \say{overdriving} the joint. For instance, if the finger makes contact when the MCP joint is at $q\degree$, instead the position demand can be set to $q + \Delta_\mathrm{MCP}\degree$.
By doing so, the finger maintains a stable contact with the environment at a similar force magnitude even when the specific shape of the contact surface is unknown.
The right side of Fig.\ref{fig:finger}A shows the normal and shear force profile of the slide back motion under $\Delta_\mathrm{MCP}= 0,\ 7.5,\ 15$mm for two repeats. 
The force magnitude increases almost linearly with the additional $\Delta_\mathrm{MCP}$. 
This plot also shows the repeatability of the soft finger, as two repeats of the same action are near identical (an average RMSE of $3.6 \pm4.3$g across all experiments).

\subsubsection{Human motion comparison}
Through the compliant MCP joint, the ADAPT Hand finger motion shows a kinematic resemblance to that a human finger at the joint displacement level. 
For the slide back and front motions, a similar motion was performed by a human whilst visually recording its pose.

Fig.\ref{fig:finger}B shows the joint evolution of the finger during the two sliding motions for a human, the robot with a soft MCP, and the robot with a rigid MCP. 
In both motions, the soft finger is most similar to the human motion. 
For the slide back motion, the difference is most notable in the relative motion of the MCP and PIP joints. The two joint angles are near consistent on the rigid finger, implying that all the motion is through the DIP joint (the image above the plot shows this in effect).
For the slide front motion, the soft finger shows even higher similarity with the human while the rigid finger barely moves. The slide front motion inherently relies on the MCP joint to flex as the PIP/DIP joints are extending, which the rigid finger cannot achieve without extra waypoints.

Although the results shown through this comparison are only indicative of the similarities between the human and robot finger motions, it motivates the necessity of a compliant MCP joint in the finger structure to provide a natural motion of the PIP/DIP joints while maintaining contact.

% Conforming to the environment.
\subsubsection{Measuring robustness through finger compliance}

In Fig.\ref{fig:finger}A and B we demonstrate human-like sliding motions with minimal planning (only three waypoints) by leveraging the compliant MCP joint.
The robustness (behavior invariance to changes in the environment) is measured by comparing task performances between the soft and rigid fingers while varying the environment. 

% FINGER SLIDING ON PLATE
Firstly, we consider the consistency of the forces generated in the two sliding motions. 
While displacing the finger and the surface in the $z$ and $\theta$ directions (see Fig.\ref{fig:finger}C top), the variability in applied forces for the rigid and soft fingers is measured.

The scatter plots in Fig.\ref{fig:finger}C compare the normal and shear forces ($F_V$, $F_H$) for the two sling motions between the soft and rigid fingers for combinations of $\Delta_z$ and $\Delta_\theta$. 
The MCP overdrive ($\Delta_\mathrm{MCP}$) on the soft finger is chosen to match the maximum force applied to the rigid finger when $\Delta_z,\ \Delta_\theta$=0 as shown by the opaque scatter point. 
The error bars show the standard deviation of maximum force recorded, where the soft finger is on average 2.4 times lower. 
The lower variability in interaction forces implies the soft finger is less influenced (i.e.: more robust) by position variability.
% This quantifies at the lowest level of interaction the soft finger is more robust to environmental uncertainty. 

% KNOB TURNING 
We can also return to the knob-turning task to evaluate the robustness when actuating the fingers. 
Here the turn angle of the knob is measured while the size of the knob and robot displacements are varied (see Fig.\ref{fig:finger}D top). 
For the soft finger, this motion can be is achieved with only three position waypoints reusing the slide back and front motions for the thumb and middle finger respectively. 
If the same waypoints are executed on the rigid finger, the robot is damaged from high forces. 
Instead, for the rigid finger, precise motion planning (15 waypoints) is necessary to describe the path specifically around the knob.
The waypoints needed for this task are shown in the timeseries graph in Fig.\ref{fig:finger}D, which highlights how the compliance in the finger simplifies control. 

% Using these controllers the adaptive performance can also be compared. 
The performance of the rigid and soft fingers is summarized in the scatter plot (Fig.\ref{fig:finger}D bottom) for three environmental settings: default settings (when the motion was programmed), reduced knob diameter, and displacement of the knob location by 1cm. 
Overall the soft finger is higher performing. As the rigid finger requires waypoints that follow the contour of the knob,  the interaction forces can vary largely compared to the soft finger, leading to inconsistent contact force application and thus lower turn angle. 
For example, with a smaller knob the turn angle increases in the soft finger (since the same turn distance covers a larger angle for a smaller $d$) by 10$\degree$, the turn angle for the rigid finger reduces by 8$\degree$.
Moreover, when the knob is displaced, the rigid finger is damaged from the high reaction forces.

% FINGER GAITING
In the finger gaiting task, the number of completed gaits until failure and the average holding force $F_\mathrm{Hold}$ is recorded while the width of the block $w$ is varied (Fig.\ref{fig:finger}E top).
Similar to knob-turning, a separate trajectory for the rigid finger was programmed which carefully tracks the contact with the block. Unlike the knob turning, the number of waypoints remain identical, but a high precision is still required for the MCP joint angle to follow the width of the block. 
The raw measurements for the completed gaits and $F_\mathrm{Hold}$ for the three $w$ settings are shown in Fig.\ref{fig:finger}E as two performance metrics. 
For both metrics, soft finger has a higher consistency as $w$ changes in comparison to the rigid finger. This is a direct extension of the result in Fig.\ref{fig:finger}C, as the interaction forces are consistent under uncertainty in displacement, resulting in a more robust behavior.

\subsubsection{In-hand cube re-orientation}
Throughout Fig.\ref{fig:finger}A-E, there are clear benefits to the motions, but they are simple. %although the benefits of the centimeter scale pose adaptation to generate robust interactions with simple waypoints planning are shown, the motion executed are elementary. 
While using the same waypoint planning methodology, we can demonstrate a more complex cube re-orientation task (see Video S1).

Fig.\ref{fig:finger}F captures key frames of the robot using its fingers, thumb, and palm to continuously re-orient a cube with a total of 12 waypoints (number of waypoints marked at every frame). 
While the entire motion is complex, it can be formed by combining the simpler motions shown in Fig.\ref{fig:finger}C-E.
The same programmed motion is repeated 20 times for cubes with three different sizes, with 100\% success rate of the small and medium cubes and 90\% for the large cube.

%%%%%%%%%%%%%%%%%%%%%%%%%%%%%%%%%%%%%%%%%%%%%%%%%%%%%%%%%%%%%%%%%%%%%%%%%%%%%%%%%%%%%%%%%%%%%%%
%%%  Sub-Section: Robustness
%%%%%%%%%%%%%%%%%%%%%%%%%%%%%%%%%%%%%%%%%%%%%%%%%%%%%%%%%%%%%%%%%%%%%%%%%%%%%%%%%%%%%%%%%%%%%%%
\subsection{Systematic pick-and-place robustness assessment of the ADAPT Hand}

\begin{figure}[tb]
    \centering
    \includegraphics[width=0.8\linewidth]{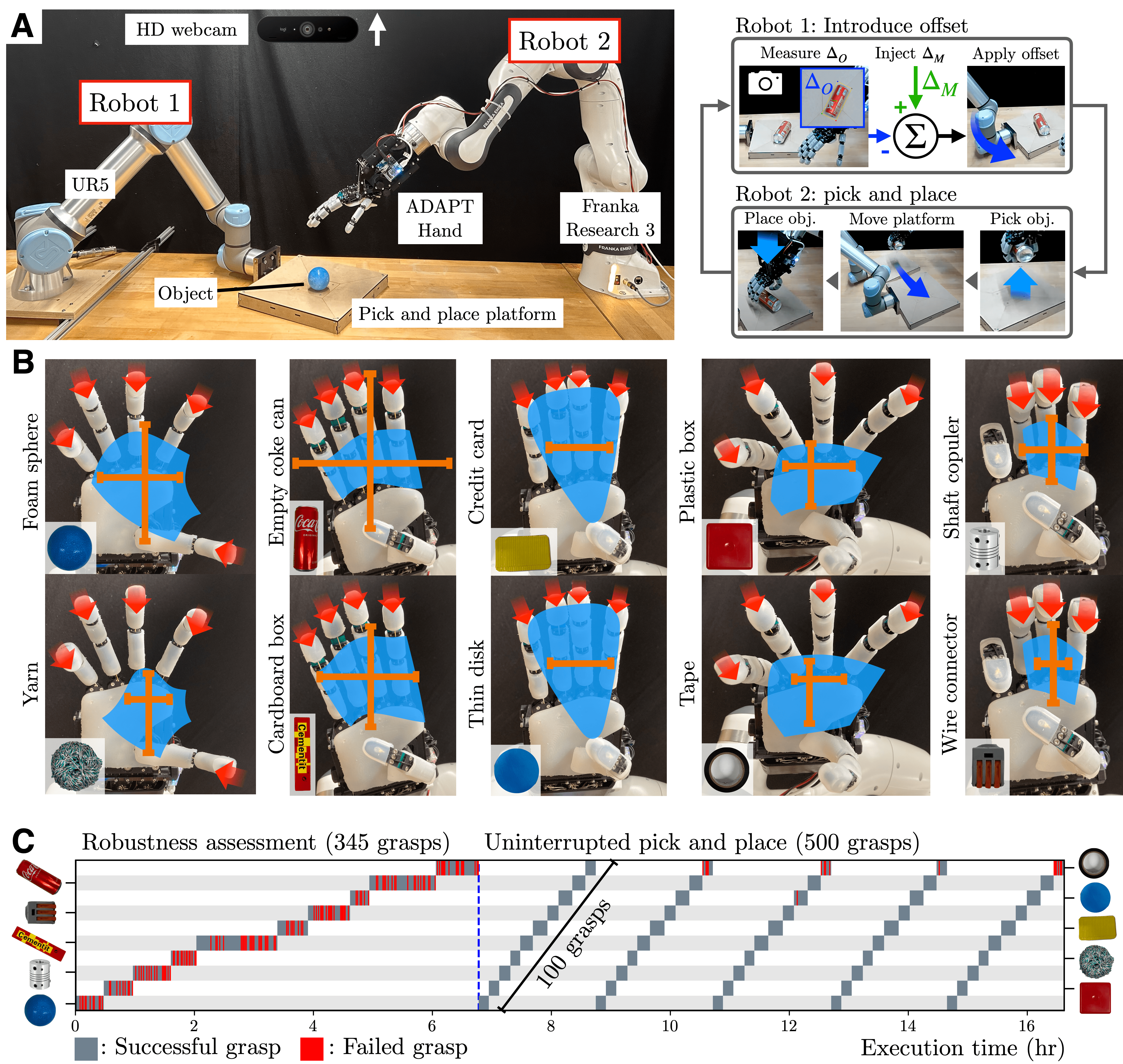}
    \caption{A) Robotic setup to conduct large quantities of automatic pick-and-place experiments while controlling the displacement of the object. B) Measured limits on object displacement (orange axis), estimated geometric limits based on object size and hand closure motion. C) Success and failed grasps throughout the two experiments: robustness assessment and uninterrupted pick-and-place totalling 845 grasps.}
    \label{fig:robust}
\end{figure}

By introducing compliance into the robot skin and finger we have shown low level measurements of improvements in the robustness and stability for interactions that require individual or multiple fingers. 
Extending to include all five digits, we now explore how this compliance influences a fundamental manipulation task, pick-and-place.  
Specifically we want to explore the 
robustness/repeatability of the hardware design through many 100s of picks, and provide quantifiable measurements for the robustness of the open-loop pick-and-place performance.
To do this, we evaluate the robots picking success as an object is iteratively displaced further away from the central point of the palm in the x and y direction.  We compare this to an estimated theoretical limit defined purely geometrically (see Fig.\ref{suppfig:feasible}).

To systematically perform this large scale evaluation of robustness, a robotic system surrounding the hand is developed to automatically perform pick-and-place tasks. This comprises of a secondary arm with a movable plate and an overhead camera (see Fig.\ref{fig:robust}A).
The process begins with Robot 1 introducing an arbitrary offset $\Delta_M$ as defined manually. To do this, the system first identifies existing displacement of the object from the center of the place: $\Delta_O$. By moving the plate by $\Delta_M - \Delta_O$, the object is now displaced by $\Delta_M$ from the plate center.
Robot 2 with the ADAPT Hand can then execute a open-loop pick-and-place motion. 
This process is outlined in Fig.\ref{fig:robust}A and can be performed continuously for many hours with minimal intervention.

\subsubsection{Measuring open-loop robustness}

Using this setup, we experimentally measure the ADAPT Hand's robustness to displacement as compared to the estimated theoretical upper limit. 
First, five objects with different geometries were chosen, and used to program a separate grasp for each object mimicking how a human would grasp.  
Then, five extra \say{unseen} objects but with similar geometries were added to the object set. 
The \say{tested} and \say{unseen} objects are labeled in Fig.\ref{fig:robust}B. 
For each objects, $\Delta_M$ was varied in the horizontal and vertical directions independently until two consecutive failed grasps were recorded.

Fig.\ref{fig:robust}B shows ten pre-grasp poses of the hand overlayed with results of measured and approximate theoretical limits of possible object displacements. 
The blue shaded area indicate an estimated geometric theoretical limit. This is defined as when the center of mass of each object could be placed within the area spanned by the motion of the fingers without colliding with the fingertips at the start of the grasp (see Fig.\ref{suppfig:feasible}).
The measured limits on the vertical and horizontal directions are indicated by the orange error bars.

Surprisingly, the hand is close to or exceeds the theoretical geometric limits of open-loop pick-and-place capabilities, despite the grasps are programmed with approximate motions and half the objects have never been evaluated.
When the objects are displaced at the millimeter scale, the contact stability from the skin compliance can maintain a similar grasp configuration (as the programmed one). 
At larger centimeter scale displacements, the effect of the finger pose adaptation is greater. The same objects under the same motion can be held by different fingers, while still being able to hold the object stably (examples of such grasps are shown in Fig.\ref{suppfig:grasping}).
Disregarding the measurements from the \say{Empty coke can} (since it is a clear outlier which skews results), the ratio of the measured over the theoretical limit on the vertical and horizontal axis are 1.2$\pm$0.2 and 0.8$\pm$0.2. 
On average, the vertical robustness exceeds the theoretical limit because even when objects collide with fingertips, the grasp is successful (an assumption use to derive the theoretical limit). 
The horizontal robustness is lower than the theoretical limit, since at large deviations the forces from the finger are too focused on one side forcing the object out of the hand. 
For the coke can, the measured limits far exceeds the theoretical because of the round shape (grasp is still successful even placed at the fingertips), and the extremely lightweight nature (can grasp just at the ends).

\subsubsection{Extended period of operation}
\label{sec:extended_opeartion}
Using the continuous test capabilities of the robotic system in Fig.\ref{fig:robust}A  the robustness and repeatability of the ADAPT Hand robot design can be evaluated for a large, uninterrupted trial.
The robot system completed 500 grasps with $\Delta_M$=0 with a total success rate of 97\%. During this operation, the robot's hardware and software system was fully untouched. 
In fact, the majority of the failures (15/16) are from grasping the tape.
This is reasonable in hindsight given the zero position of the tape was already close to the edge of the robustness limits (the point of intersection of the orange error bars in Fig.\ref{fig:robust}B lies near its edge).

Fig.\ref{fig:robust}C shows a time series of all experiment performed using the pick-and-place system. Divided by the blue dotted line, the first set of experiments are for assessing the robustness described in Fig.\ref{fig:robust}B. The second set represents the uninterrupted robot operation. 
In the combined two experiments, the ADAPT Hand was actively used over 16 hours and 845 grasps without any modifications to the hardware/software. The only visible damage to the hardware was the ware/dirt on the silicone fingertips shown in Fig.\ref{suppfig:silicone}.
Video S2 shows a sped up version of this extended trial, while details to the raw video can be found in section \ref{suppsec:resource}.

%%%%%%%%%%%%%%%%%%%%%%%%%%%%%%%%%%%%%%%%%%%%%%%%%%%%%%%%%%%%%%%%%%%%%%%%%%%%%%%%%%%%%%%%%%%%%%%
%%%  Sub-Section: Wrist driven self-organization
%%%%%%%%%%%%%%%%%%%%%%%%%%%%%%%%%%%%%%%%%%%%%%%%%%%%%%%%%%%%%%%%%%%%%%%%%%%%%%%%%%%%%%%%%%%%%%%
\subsection{Self-organizing grasps}

\begin{figure}[tb]
    \centering
    \includegraphics[width=0.95\linewidth]{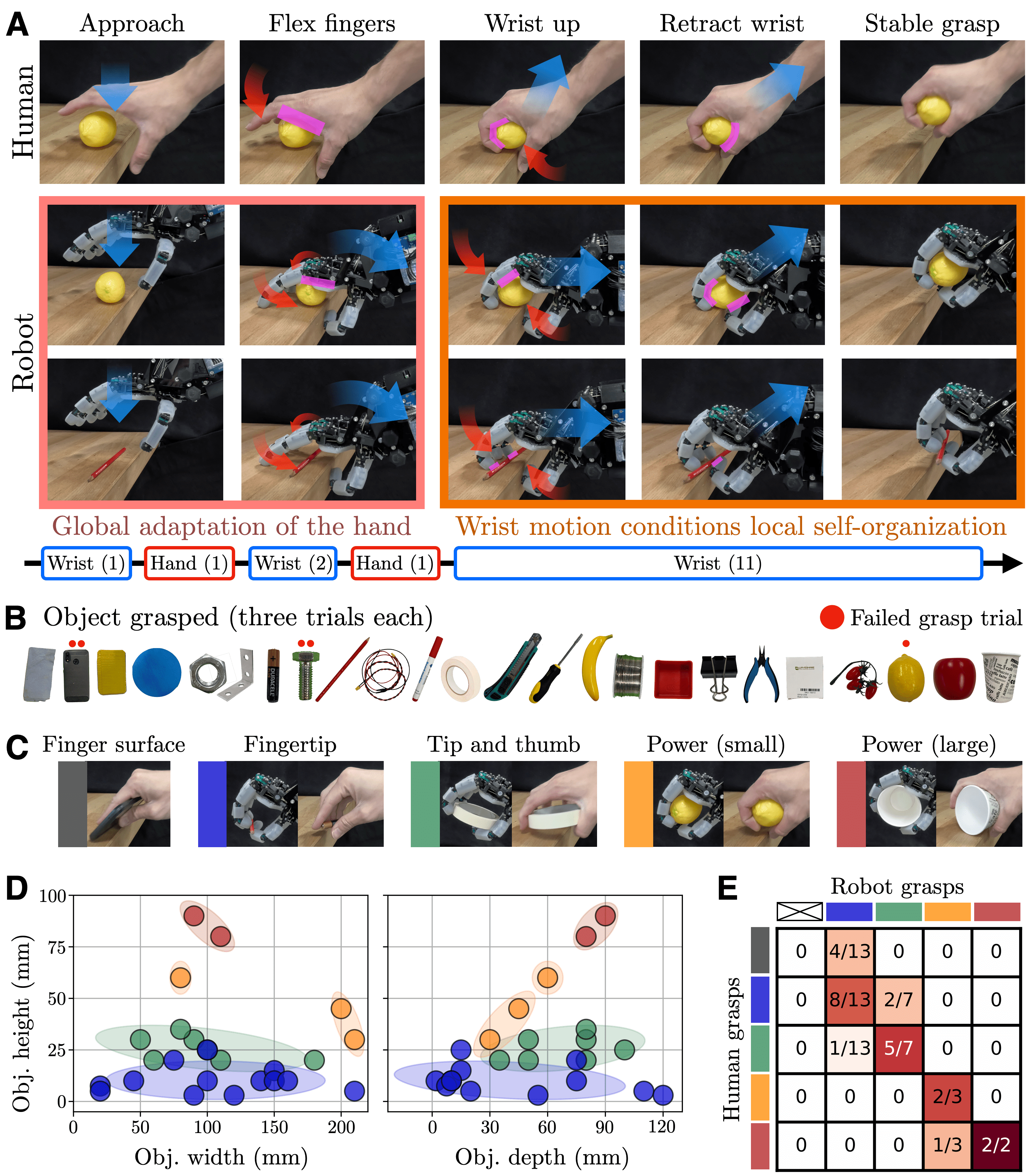}
    \caption{A) Key frames overlayed with wrist motion (blue arrow), finger motion (red arrow), and new contact regions (pink shade) for the human and robot grasping an object off the table. The number of waypoints for the wrist or hand illustrated underneath the frames. B) Objects used for this experiment with failed trials indicated by red circles above the image. C) Four grasps identified on the human and robot. D) Object geometry plotted with the color indicating the grasp type observed. E) Direct comparison between robot and human grasps for every object.}
    \label{fig:wrist}
\end{figure}

In the previous section, the focus was on the skin and finger compliance which offers robustness and stability to local interactions.
A compliant wrist enables the scale of these interactions to be expanded, for example grasping while sliding and object off a table (Fig.\ref{fig:wrist}A). 
% an action natural to humans but challenging for robots where the object must be grasped while making and breaking contact with environmental constraints.
Through open-loop trajectories of the hand and wrist which mimics human grasping of grasping objects from a table, we explore how robust this is to different objects, and also how different grasp types emerge.

When a human accomplishes this task, three behaviors within the trajectory can be identified that are invariant to the object being grasped (top row of Fig.\ref{fig:wrist}A shows an example of grasping a lemon with its key frames and the finger/wrist movements).
Firstly, the approaching motion is always the hand moving downwards until making contact, either with the table or the object.
Secondly, after the initial contact is made the fingers continue to hold its contact during the motion, either with the table or the object.
Thirdly, the grasping motion shown by the \say{Flex fingers}, \say{Wrist up}, and \say{Retract wrist} frames is a single continuous motion where the fingers and thumb curls to form a grasp as the wrist moves away from the table.

By using a wrist with human-matched compliance in addition to the skin and fingers, the observed human behaviors can be robustly replicated on the robot (as shown in the bottom two rows of Fig.\ref{fig:wrist}A).  
When approaching the table, the compliance allows for a safe and controlled contact of the fingers and/or palm as the wrist presses into the table (first two frames).
The grasping motion (the final three frames of Fig.\ref{fig:wrist}A) shows a coordinated flexing movement of fingers and thumb as the wrist moves away while constantly maintaining contact with the environment, matching with the second and third observations from the human (last three frames).
This motion relies on the distributed compliance, in particular the compliance of the wrist conditions(provides the necessary conditions) the interaction of the fingers with the environment.
During the grasping sequence, only the wrist is actively actuated (see the sequence of waypoints executed in Fig.\ref{fig:wrist}A bottom). The fingers are only actuated before the final wrist motion to from a grasp, where the movement is constrained by the table. 
The wrist movement gradually removes the constraint of the table enabling the fingers self-organize and grasp the object, while the compliance of the wrist maintains contact with the surface.

Using the same open-loop motion, 24 objects of varying geometry were chosen to be grasped from the table. 
Fig.\ref{fig:wrist}B shows each object, spanning from flat and thin objects such as a pencil, to a large and bulky objects to an apple. 
Each object was grasped three times with varying placement locations in the vicinity of the robot hand motion path, with an overall success rate of 93\% (70/75 tests) failing to grasp only three objects (phone, bolt, and lemon). 
Video S3 shows this grasping experiment for all items from two view angles.

\subsubsection{Emergence of discrete grasp types}
The robustness of the robot to grasp a wide variety of objects through identical commands can be explained through self-organization. 
% Unlike the pick-and-place in Fig.\ref{fig:robust} where the interaction with the object is at the instant when the pre-determined grasp is executed, in this experiment the grasp results as an emergence of the continuous contact and passive adaptation of the fingers. 
% The emergence of grasps means the robot holds each object differently in the most stable configuration, hence the robustness. 
Consider the two robot motions given in Fig.\ref{fig:wrist}A. In the top row, the lemon is a large object where the robot makes multiple contacts throughout the motion (see pink highlights). On the bottom row, the robot only makes contact with a pencil just when the grasp happens. The resultant grasp of the two objects are also different: the lemon is held with a power grasp while the pencil is held only with the fingertips, which is appropriate if a human were to grasp the two objects. 

For all the objects, the grasp motions were analyzed for the robot and the human, and categorized to five distinct grasp types shown in Fig.\ref{fig:wrist}C. 
The grasp types are categorized based on the hand posture and which parts of the hand interact with the object which relate with the grasping taxonomies (see section \ref{sec:meth_grasp_categorization}). Video S3 shows the four grasp types identified for the robot. 

The scatter plot in Fig.\ref{fig:wrist}C shows the object geometry (length, width, height) clustered by the observed grasp types of the robot.
The clustering shows the robot hand self-organizes to a discrete set of grasp types based on the object's geometry.
For flat or small objects the grasp is using its fingertips (blue cluster). As the height increases, the robot starts to use its thumb (green cluster), slowly transitioning into a form of power grasp (yellow, red clusters). 

Like the example in Fig.\ref{fig:wrist}A, the robot grasp types observed match closely with a human. 
It is know that a human will vary the grasping strategy based on the object geometry (and other factors too)\cite{feix2014analysis}, which is indeed observed when asked to grasp the objects used in this experiment. 

The matrix in Fig.\ref{fig:wrist}C directly compares grasps observed for the robot and human. 
Categorically the human exhibits one extra grasp type, the finger surface grasp where all fingers a kept straight, which is impossible for the robot as the programmed motion curls the PIP/DIP finger joints.
Despite this mismatch of four versus five grasps, the strong correspondence between the two clear by 68\% (17/24 objects) lying on the leading diagonal, and all other objects on directly adjacent cells. 
Assuming the human is choosing the optimally stable grasp through prior knowledge, the robot having a biomimicry of the kinematics and distributed stiffness is able to self-organize to choose the same. 

%%%%%%%%%%%%%%%%%%%%%%%%%%%%%%%%%%%%%%%%%%%%%%%%%%%%%%%%%%%%%%%%%%%%%%%%%%%%%%%%%%%%%%%%%%%%%%%
%%%  SECTION: Discussion / Conclusion
%%%%%%%%%%%%%%%%%%%%%%%%%%%%%%%%%%%%%%%%%%%%%%%%%%%%%%%%%%%%%%%%%%%%%%%%%%%%%%%%%%%%%%%%%%%%%%%
\section{Discussion}

%% Summary of work
In this work, we present an anthropomorphic robot hand, the ADAPT Hand, designed with a biomimetic distribution of compliance across different lengths scales in the skin, finger, and wrist. 
Starting from low level interactions of the skin and finger levels, we show the presence of a compliant skin and MCP joint on the finger leads to higher performance and robust to environmental uncertainty across three of tasks. 
Through the ADAPT Hand's tunable compliance, a direct comparison between a compliant and rigid robotic hardware is made.
Expanding the task to include all five digits, the robustness within the hand is measured to lie close to an estimated theoretical limit for a pick-and-place task.
Using the same measurement setup, the robustness and repeatability of the hardware platform itself is shown through a damage-less execution of over 800 grasps and 15 hours of operation.
Finally, a compliant wrist motion is introduced to grasp 24 objects through an interaction-rich motion across a tabletop surface. 
This motion which fully combines the distribution of compliance across the robot body, where a wrist motion which conditions the passive pose adaptation of the fingers while contact stability from the skin enhances the grasp.
The resultant discrete grasp types are self-organized based on the object geometry, similar to that of a human. 
Not only, a comparison between human and robot grasp shows 68\% of grasps are directly matching.
Overall, we demonstrate an physical intelligence approach towards anthropomorphic robot design which considers the interaction and motion at all the lengths scales of robotic manipulation. 
The extreme robustness given simplest form of actuation control input culminating in the self-organizing grasps.

Although the proposed concept realized by the ADAPT Hand show remarkable performance, there are clear limitations to be addressed in the future. 
At the heart of this work is the distributed compliance, which can negatively affect certain tasks.
For instance the soft skin and finger limits the ability to exert high forces in one direction. 
This means strongly pinching an object or button pressing tasks are challenging. 
When using the full hand for manipulation, the passive adaptation is simply a result of force equilibrium working in a desired way. This means, when the external forces are too large (e.g.: an object that is too heavy such as the phone in Fig.\ref{fig:wrist}B), the execution fails. 
The same is true for balancing forces between multiple fingers and a held object. Since the fingers will passively adapt, tasks such as controlled object re-orientation is difficult.

Sensor feedback and control is one obvious next step to tackle these limitations. 
This can mean sensing of the joint torques/tendon forces as well as tactile sensing to span the skin surface.
However, the control method using the sensor feedback is not straightforward for this robot. 
This is because the manipulation capabilities of this hand is driven through the emergent behaviours which arise from the interaction between the compliant robot and the environment. 
Although the actuation signals are position commands, each waypoint is programmed with a higher level intention such as \say{continue to apply force downwards} or \say{move roughly in this direction} instead of reaching a particular position or force.
The sensory-motor control should hence not follow some explicit trajectory, but should rather complement the emergent of behavior. 
This could be achieved through combining high level \say{intention} in the form of the open-loop waypoints and a controller to regulate contact forces, similar to a shared-control scheme used in\cite{khadivar2022emg} used to stabilize a noisy EMG signal.
Alternatively a bio-inspired sensory-motor control, such as central pattern generators(CPGs) used to transition between walking and swimming gaits for a salamander robot\cite{ijspeert2007swimming}, can lead to emergent efficient behaviors.
While the concept of CPGs do not directly apply to manipulation, a sensory-motor coordination method that drives emergent behaviours\cite{pfeifer2007self} is necessary to maintain and extend the robust interactions.

The ADAPT Hand as a hardware platform itself also has room for improvement. 
One direction is to increase the level of biomimicry of the human hand. The current hand although is bio-inspired, clear differences are present especially in the joint placement and the skin/flesh distribution. For example, there is no material covering the MCP joints. Consequentially, multiple grasp formations do not translate well from a human to the robot. 
Another limitation is the lack of a wrist just below the hand. While the 7dof robot arm can orient the hand in any angle, in practice the arm must move a large displacement to resemble small wrist motions seen in a human. 
Likewise to the hand itself, increasing the biomimicry in the arm kinematics can enable more complex human motions to be realised by the robot. 
Not only could this simplify manual programming or teleoperation by a human, but one could imagine mapping motions learnt from the abundance of videos of humans to be applied to this biomimetic robotic hand-arm structure.

%%%%%%%%%%%%%%%%%%%%%%%%%%%%%%%%%%%%%%%%%%%%%%%%%%%%%%%%%%%%%%%%%%%%%%%%%%%%%%%%%%%%%%%%%%%%%%%
%%%  SECTION: Methods
%%%%%%%%%%%%%%%%%%%%%%%%%%%%%%%%%%%%%%%%%%%%%%%%%%%%%%%%%%%%%%%%%%%%%%%%%%%%%%%%%%%%%%%%%%%%%%%
\section{Methods}

\begin{figure}[tb]
    \centering
    \includegraphics[width=0.95\linewidth]{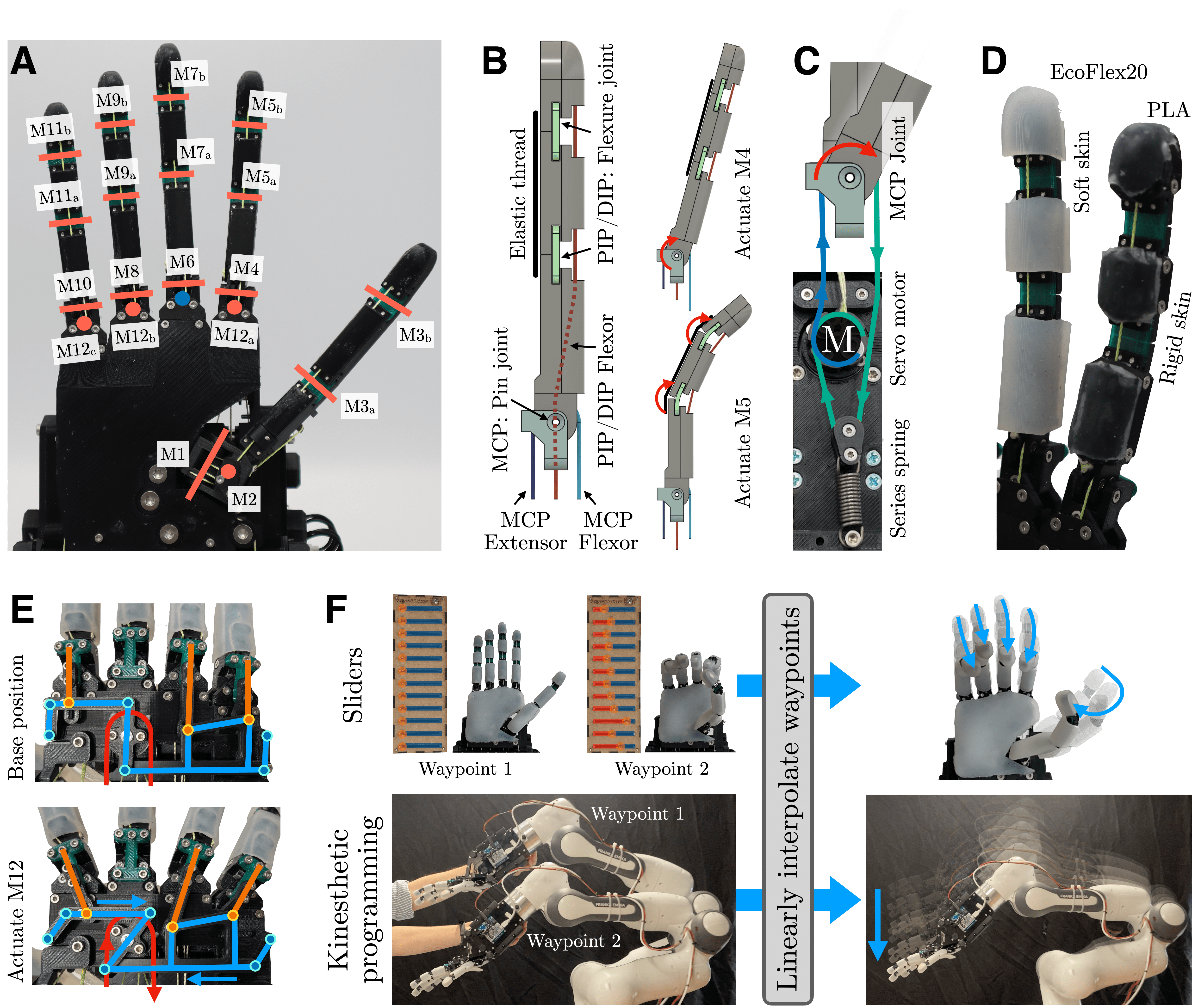}
    \caption{A) Actuated, underactuated, and passive joints for the ADAPT Hand. B) Finger design and tendon routing. C) Series elastic actuation design for the the MCP joint. D) Soft and rigid skin designs. E) Abduction/adduction linkage mechanism. F) Waypoint recording and replay method used to program the robot motion.}
    \label{fig:robot}
\end{figure}

%%%%%%%%%%%%%%%%%%%%%%%%%%%%%%%%%%%%%%%%%%%%%%%%%%%%%%%%%%%%%%%%%%%%%%%%%%%%%%%%%%%%%%%%%%%%%%%
%%%  Sub-Section: ADAPT Robustness
%%%%%%%%%%%%%%%%%%%%%%%%%%%%%%%%%%%%%%%%%%%%%%%%%%%%%%%%%%%%%%%%%%%%%%%%%%%%%%%%%%%%%%%%%%%%%%%
\subsection{ADAPT Hand Hardware}
\label{sec:meth_adapthand}
The ADAPT Hand is a custom designed and fabricated anthropomorphic robot hand, with its four fingers and thumb having dimensions similar to an adult human. 
The entire hand is fabricated from commercially available 3D printed Polylactic acid(PLA) and Thermoplastic Polyurethane(TPU) with shore hardness 98A/65D, and uses tendons (cables) for its actuation. Fig.\ref{fig:robot}A shows the ADAPT Hand without the skin, showing the joint kinematics and actuation scheme. 
A total of 12 servo motors (Dynamixel XM430-W210-R) located beneath the hand actuates the 20 joints (four joints per finger/thumb). 
Having less actuators than joints means certain joints are underactuated (shown by subscripts a, b, c in Fig.\ref{fig:robot}A) or passive (\say{P} in Fig.\ref{fig:robot}A). Details on the CAD model is presented in section \ref{suppsec:resource}. 

\subsubsection{Finger design}
The finger design shown in Fig.\ref{fig:robot}B is a key feature of the ADAPT Hand, where the same design is used across the five digits.
One actuator controls the pin-jointed MCP joint with antagonistic tendons (the thumb has an extra antagonistic joint, but rotate 90 degrees to make the 2DoF CMC joint). 
The PIP/DIP flexure joints which are coupled by a single flexor tendon is actuated by another motor. 
The pin joint in the MCP joint allows the routing of the PIP/DIP flexor to pass through the center of rotation, fully decoupling the two axis of actuation. 
The extension forces for the PIP/DIP come from the combined effect of the TPU flexure joint and the elastic thread, placed on the backside of the finger (shown by a black line in Fig.\ref{fig:robot}B).

The compliance at the finger level is generated by a series elastic MCP joint which is achieved by routing the MCP flexor tendon around a pulley connected to an extension spring (shown in Fig.\ref{fig:robot}C). 
Replacing or removing the series spring can change the finger compliance (as in Fig.\ref{fig:bio-inspiration}B).

\subsubsection{MCP abduction-adduction motion}
A notable feature of the mechanical design of the ADAPT Hand is the series-elastic linkage mechanism driving the abduction-adduction axis of the MCP joints for the index, ring, and little fingers. 
The mechanism and its open and closed states are shown in Fig.\ref{fig:robot}E. The linkage mechanism connects to the MCP pin joint with a series elastic TPU material, making the MCP joint compliant in 2 axes.
Having an actuated spread axis of the MCP joints increases the workspace (thus the capability of the hand) such as holding large objects.

\subsubsection{Dexterity of the ADAPT Hand}
Combining all the mechanisms of the fingers, the ADAPT Hand can be actuated to produce dexterous motions. Starting from the zero position (all fingers straight), the hand can be actuated to produce all 33 grasp taxonomies\cite{feix2015grasp}. 
Likewise, the hand is able to complete all 10 postures on the Kapandji test. The results for both tests are shown in Fig.\ref{suppfig:taxonomy}. 

\subsubsection{Skin design}
A modular skin fully covers one side of the ADAPT Hand. 
Fig.\ref{fig:robot}D shows the index and middle finger equipped with a rigid and soft skin respectively. The skins are identical in their geometry and is approximately 3mm offset from the \say{bones}. 
The soft skin is fabricated from cast EcoFlex20 and the rigid skin is 3D printed PLA. A thin ($\approx0.5$mm) layer of EcoFlex20 is glued on the surface of the rigid skin to maintain the surface friction property.

%%%%%%%%%%%%%%%%%%%%%%%%%%%%%%%%%%%%%%%%%%%%%%%%%%%%%%%%%%%%%%%%%%%%%%%%%%%%%%%%%%%%%%%%%%%%%%%
%%%  Sub-Section: ADAPT motion programming
%%%%%%%%%%%%%%%%%%%%%%%%%%%%%%%%%%%%%%%%%%%%%%%%%%%%%%%%%%%%%%%%%%%%%%%%%%%%%%%%%%%%%%%%%%%%%%%
\subsection{ADAPT Hand motion programming}
\label{sec:meth_teaching}
In all experiments, the ADAPT Hand (including the robotic arm) operates through manually programmed open-loop motions. 
For both the hand and arm, a series of manually determined key waypoints are recorded to then be played back (see Fig.\ref{fig:robot}F).
The source code details are presented in section \ref{suppsec:resource}, with the software system integration and hardware interfacing which allows the programming and replay of waypoints on the hand and arm described in Fig.\ref{suppfig:diagram}.

\subsubsection{Programming the hand}
The ADAPT Hand is controlled by directly commanding the tendon displacements for each actuator (which is proportional to the motor angle). 
To simplify the procedure to manually record waypoints, the hand (fingers) is operated using a custom built \say{signal mixer box} with 12 linear potentiometers which map to the position of each actuator. 
As illustrated in the top left of Fig.\ref{fig:robot}F, two waypoints can be simply defined by varying the linear potentiometer positions. 
Being an interactive device, the mixer box allows for quickly programming motions while having full control over the tenon positions. 

Once a set of one or more waypoints (slider position) are recorded, the robot can smoothly move between the waypoints as in the top right of Fig.\ref{fig:robot}F. 
% Fig.\ref{suppfig:diagram} shows the software system integration and hardware interfacing which allows the programming and replay of the waypoints. 

\subsubsection{Programming the arm}
The Franka Research 3 robot arm is controlled using a gravity compensated impedance control introduced in \cite{franzese2021ilosa}, where the end effector 6 dof pose and corresponding stiffness can be commanded. 
To program the arm, the end effector stiffness is set to zero which allows the arm to be manually moved around. 
Likewise to the hand, after recording few key poses (see bottom left of Fig.\ref{fig:robot}F), the 6 dof waypoints are interpolated and replayed. 

For motions which involve both the hand and arm waypoints are replayed sequentially, meaning the hand and arm are not actively actuated at the same time.

%%%%%%%%%%%%%%%%%%%%%%%%%%%%%%%%%%%%%%%%%%%%%%%%%%%%%%%%%%%%%%%%%%%%%%%%%%%%%%%%%%%%%%%%%%%%%%%
%%%  Sub-Section: Experimental procedure
%%%%%%%%%%%%%%%%%%%%%%%%%%%%%%%%%%%%%%%%%%%%%%%%%%%%%%%%%%%%%%%%%%%%%%%%%%%%%%%%%%%%%%%%%%%%%%%
\subsection{Experimental setup and procedure}
\label{sec:meth_exp_procedure}

\subsubsection{Measuring human compliance}
\label{sec:meth_human_measure}

The compliance (force displacement characteristics) for a human is measured by recording the reaction force on a loadcell and its Cartesian position as it is moved by a 6-axis robot arm (UR5), while the human is instructed to relax.
The measurement setup and movement directions for the skin, finger and wrist is shown in Fig.\ref{suppfig:probing}.
The forces are recorded for multiple repeats to capture the variation in the muscle activity.

\subsubsection{Measuring low-level interactions}
\label{sec:meth_low_level_exp}

\begin{figure}[tb]
    \centering
    \includegraphics[width=0.95\linewidth]{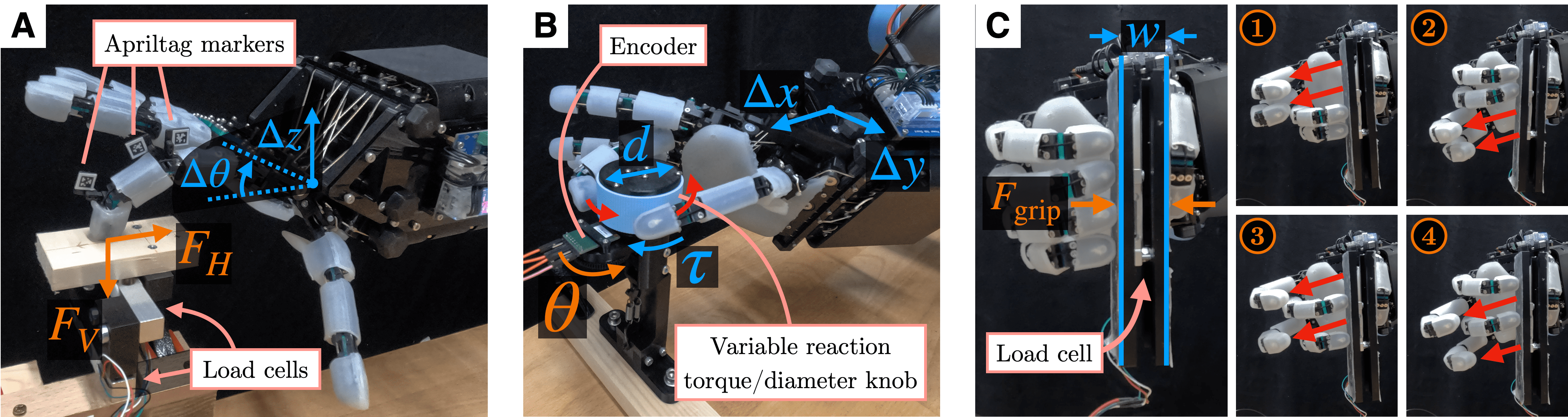}
    \caption{Experimental setup and independent (in blue) and dependent (in orange) variables for the finger sliding experiment (A), knob turning experiment (B), and the finger gaiting experiment (C).}
    \label{fig:experiment}
\end{figure}

\begin{figure}[tbh!]
    \renewcommand\thefigure{S1} 
    \centering
    \includegraphics[width=0.99\linewidth]{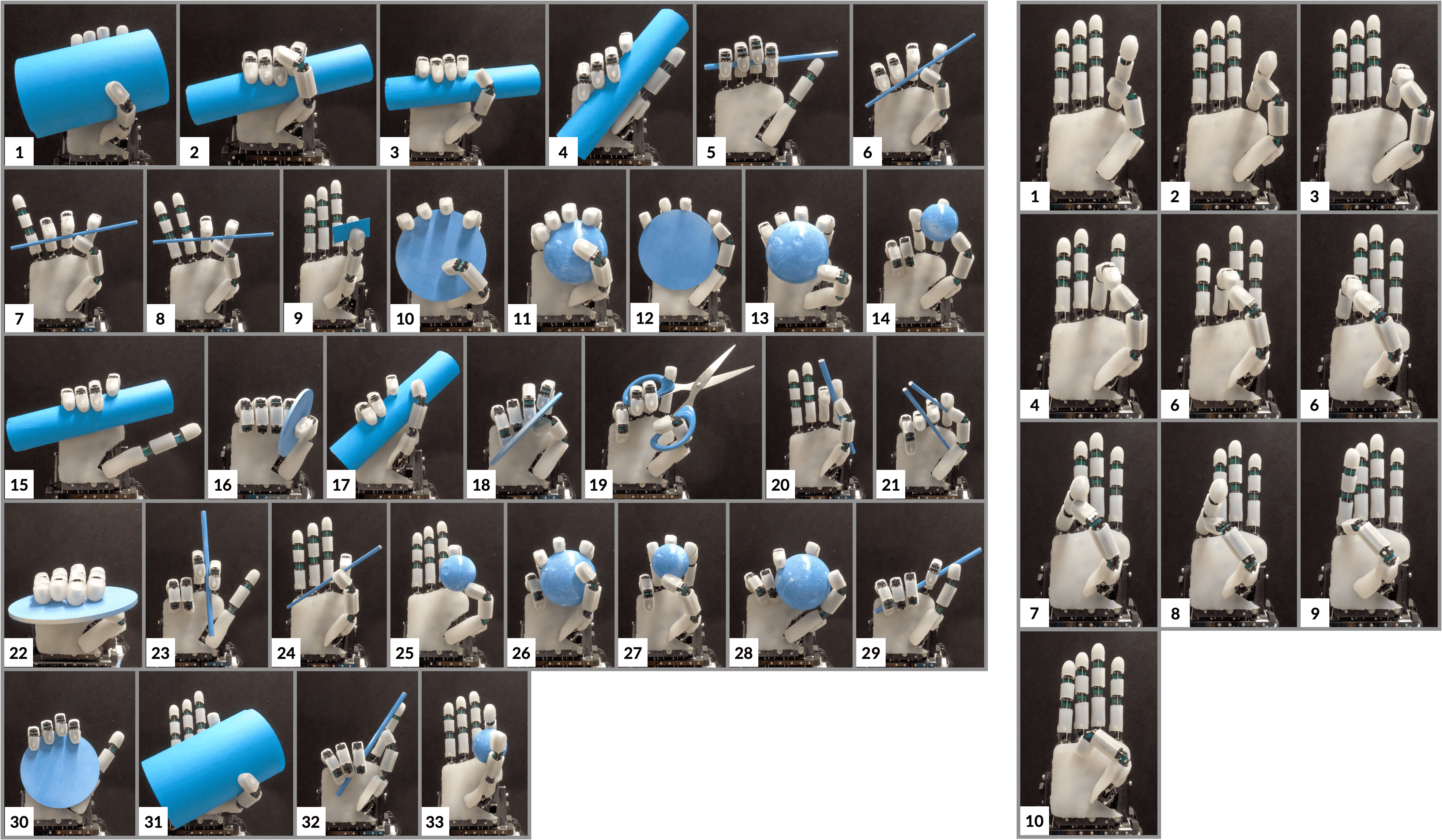}
    \caption{The ADAPT Hand showing all 33 grasp taxonomies and the 10 Kapandji score posture.}
    \label{suppfig:taxonomy}
\end{figure}

Three tasks: finger sliding, knob turning, and finger gaiting, were conducted to characterise the low level interactions (such as contact forces and kinematics) of the ADAPT Hand skin/finger with the environment. 
The experimental setup are shown in Fig.\ref{fig:experiment}A,B,C for the three tasks respectively.

\begin{figure}[tbh!]
    \renewcommand\thefigure{S2} 
    \centering
    \includegraphics[width=0.99\linewidth]{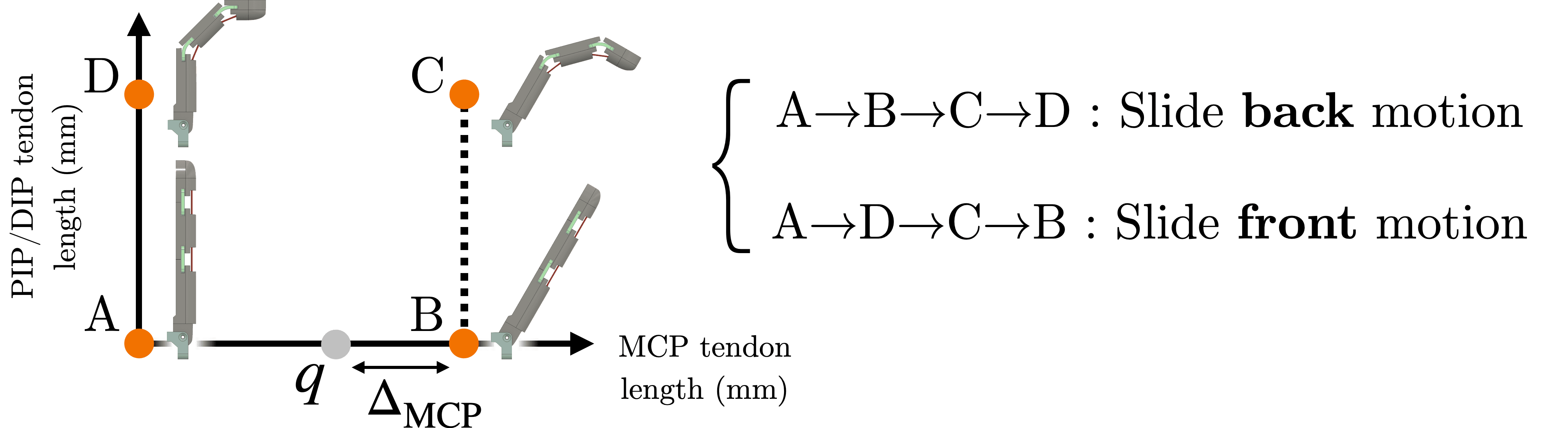}
    \caption{Visual representation of the tendon actuation space for the flexion-extension motion of a single finger. The axis show the two independent motion control for the finger. In the sliding experiment, the slide back and slide front motions are developed by moving between the four waypoints in a clockwise or counterclockwise fashion. $\Delta_\mathrm{MCP}$ shows the how much the MCP joint is overdriven.}
    \label{suppfig:finger}
\end{figure}

\begin{figure}[tbh!]
    \renewcommand\thefigure{S3} 
    \centering
    \includegraphics[width=0.99\linewidth]{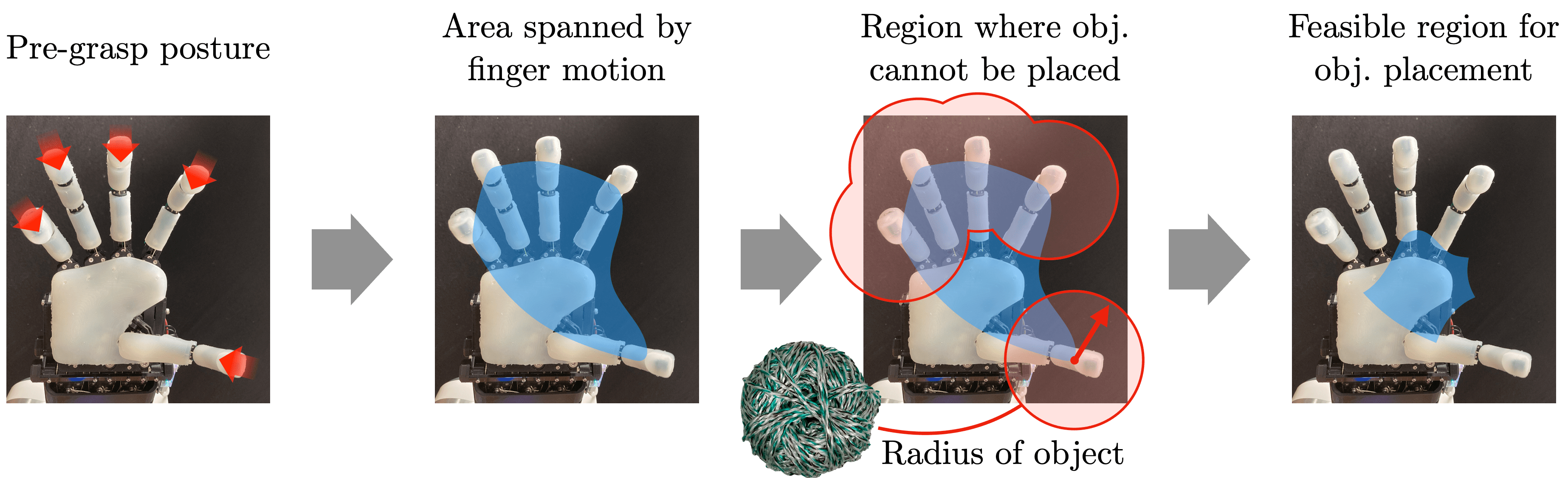}
    \caption{The flowchart used to estimate the geometrical limits for open-loop grasping an object. First, the directions of the finger/thumb motion for the grasp is noted. Then, an approximate area spanned by the finger/thumb in this grasping motion is obtained. Assuming the center of mass of the object must lie within the fingertips, an area where the object would collide with the fingertips is found. Finally, subtracting the two areas gives the estimated geometrical limits of where the center of mass of the object can be placed.}
    \label{suppfig:feasible}
\end{figure}

\begin{figure}[tbh!]
    \renewcommand\thefigure{S4} 
    \centering
    \includegraphics[width=0.99\linewidth]{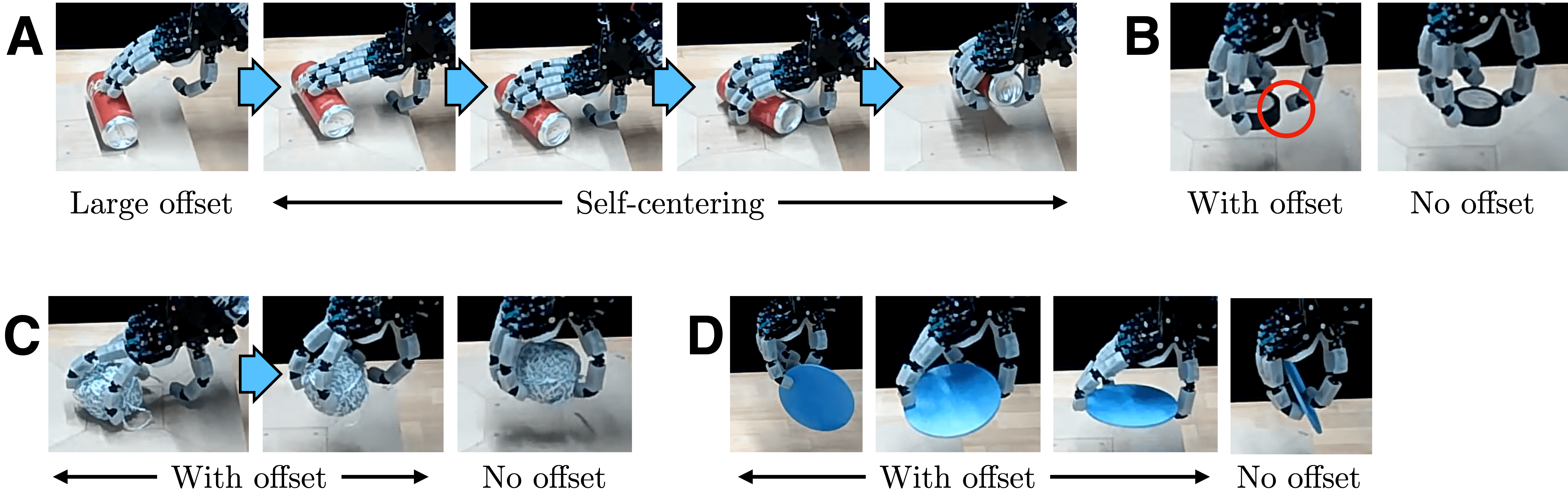}
    \caption{Four examples of the ADAPT Hand grasping objects when an artificial offset is given. 
    A: Even with a large offset, the compliance and the motion of the hand self-centers the coke can to be grasped. B, C, D: Examples where the resultant grasp differs from that of $\Delta_M=0$.}
    \label{suppfig:grasping}
\end{figure}

\begin{figure}[tbh!]
    \renewcommand\thefigure{S5} 
    \centering
    \includegraphics[width=0.65\linewidth]{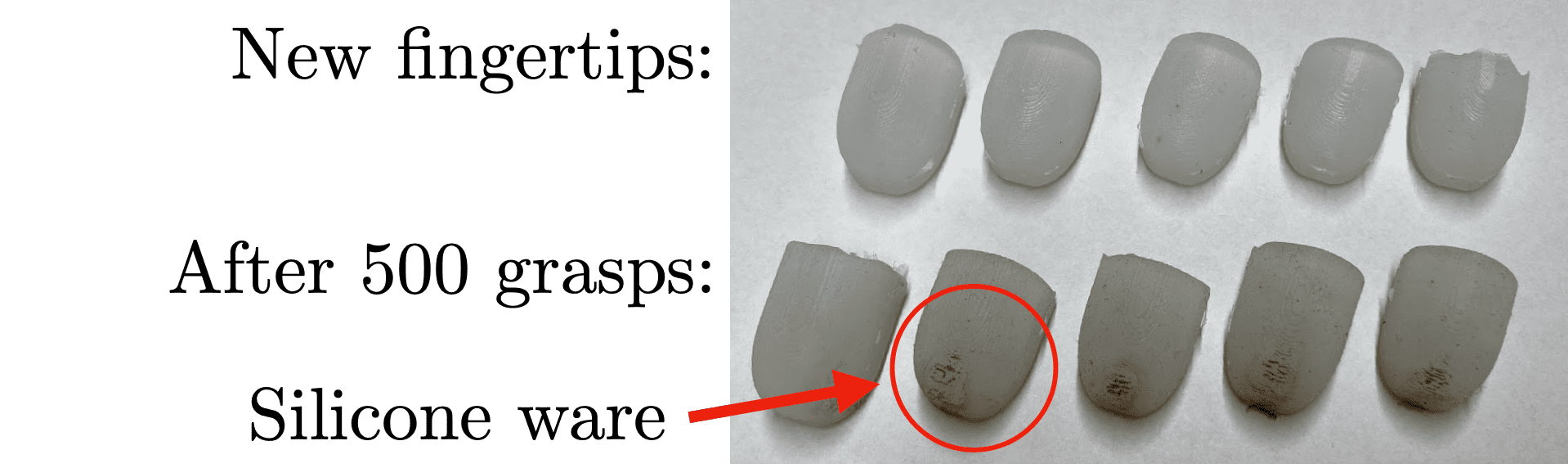}
    \caption{Photograph to show the ware and tare of the silicone fingertips before and after the extended pick and place experiment.}
    \label{suppfig:silicone}
\end{figure}

\begin{figure}[tbh!]
    \renewcommand\thefigure{S6} 
    \centering
    \includegraphics[width=0.99\linewidth]{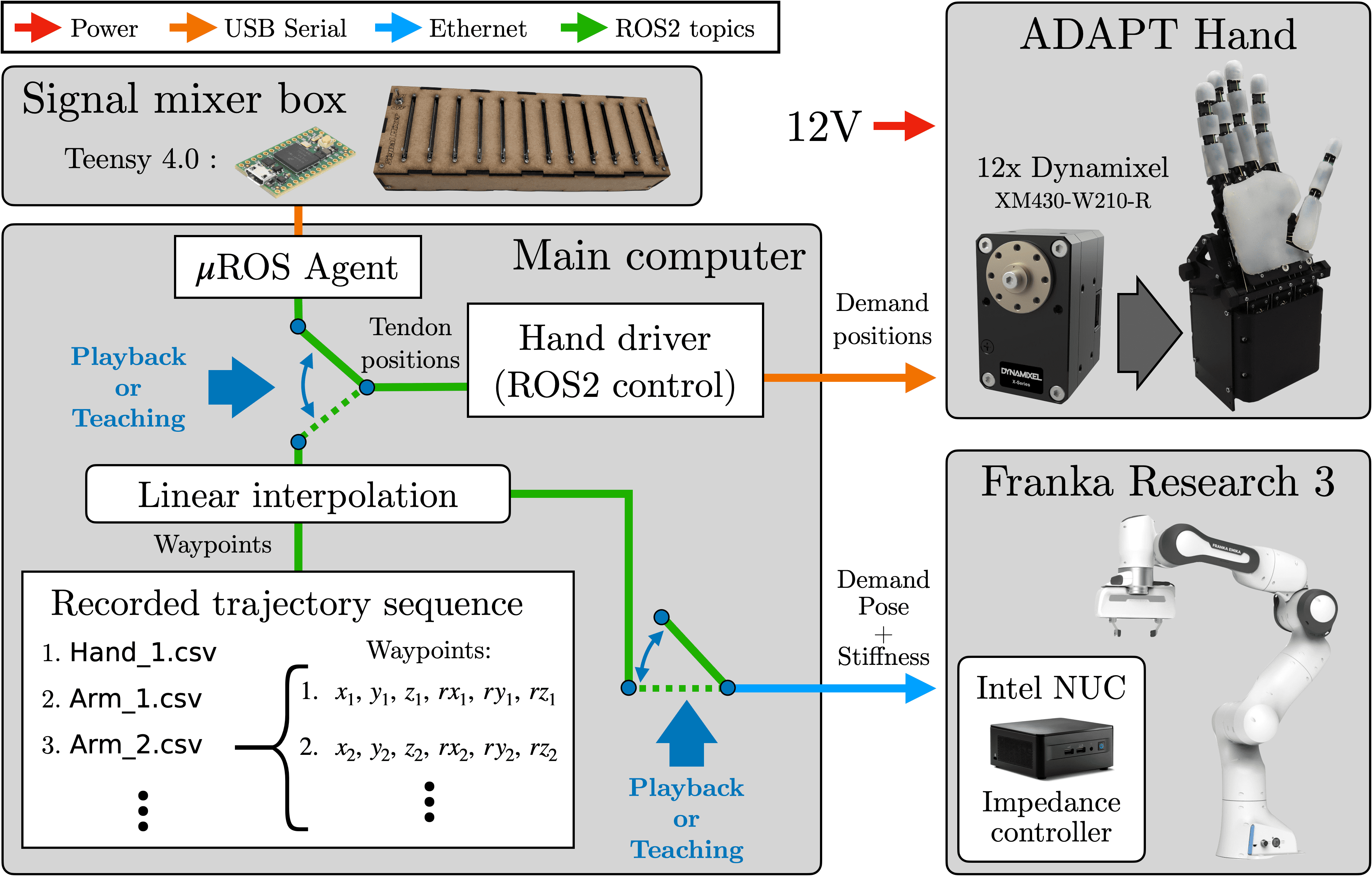}
    \caption{A high level software and system integration diagram used to control the ADAPT Hand. The heart of the control is the teaching and playback system described in section \ref{sec:meth_teaching}. When teaching (recording) different motions, the hand is actuated using a signal mixer box where the sliders correspond to demand tendon positions. No control input is given to the arm to be moved freely by hand. In the playback mode, the recorded waypoints (a sequence of csv files for the hand or the arm) will be executed sequentially either to the hand or arm.
    The full system is developed using ROS2 to interface between microcontrollers (micro-ros), custom processes, ADAPT Hand (ROS2 control driver), and the Franka impedance controller\cite{franzese2021ilosa}.}
    \label{suppfig:diagram}
\end{figure}

\begin{figure}[tbh!]
    \renewcommand\thefigure{S7} 
    \centering
    \includegraphics[width=0.99\linewidth]{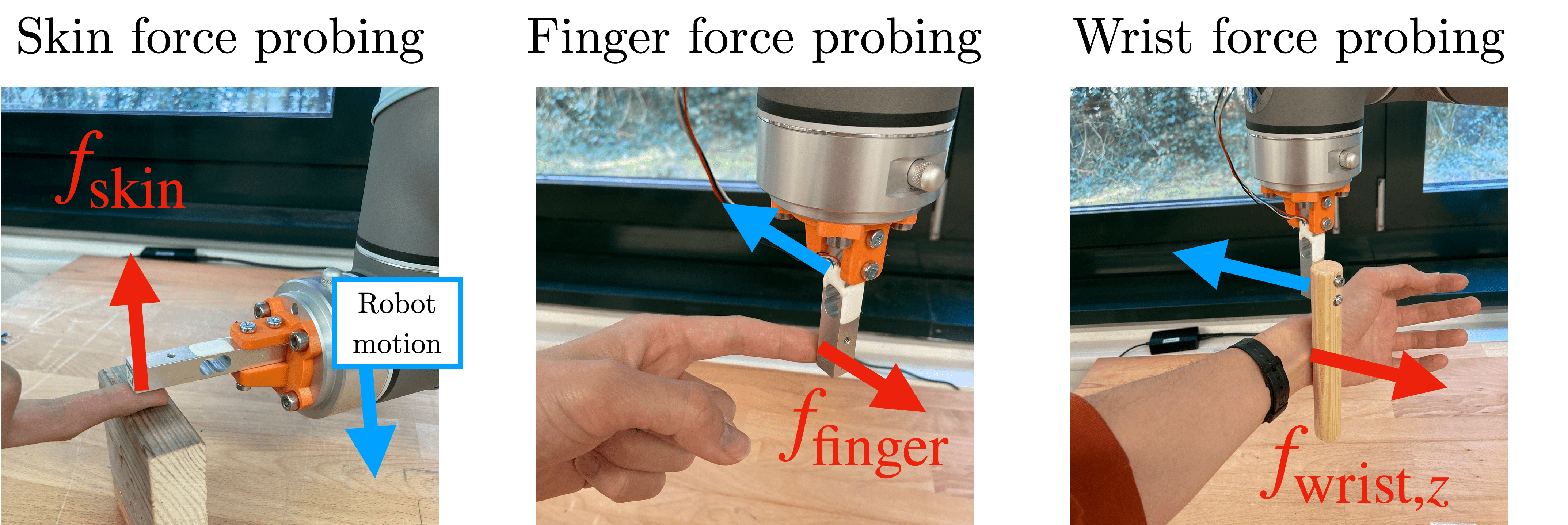}
    \caption{The measurement of human compliance (force-displacement characteristics) using a UR5 robot arm mounted with a 10kg loadcell. The robot arm is moved in different axis to while the human is instructed to relax.}
    \label{suppfig:probing}
\end{figure}

In the finger sliding task, a single finger interacts with a wooden plate through two sliding motions generated by combining a flexing motion of the MCP joint and a flexing or extending motion of the PIP/DIP joints (see Fig.\ref{suppfig:finger}). 
Fig.\ref{fig:experiment}A shows the experimental setup where the ADAPT Hand (rigid held by a UR5 arm) interacts with the wooden plate mounted above two load cells measuring the vertical ($F_V$) and horizontal ($F_H$) forces.
The finger joint angle data used in Fig.\ref{fig:finger}B were extracted by recording the April tag markers throughout the motion. 
For the two motions, five independent variables were combinatorialy tested with two repeats which are: pose offset $\Delta \theta$ ($\pm 10\deg$), $\Delta z$ ($\pm 10$mm), soft and rigid configurations for the skin and finger, and overdrive of the MCP tendon $\Delta_\mathrm{MCP}$ ($\pm 7.5$mm) (only for the soft finger).

In the knob turning task, the middle finger and thumb was used to turn a knob shown in Fig.\ref{fig:experiment}B. 
The knob turn angle $\theta$ is used to assess the performance measured by a position encoder (AMS AS5048B). 
The environment was varied in three ways: the $x$ and $y$ position offset($\pm 10$mm each), the diameter of the knob $d$($45\pm 15$mm), and the reaction torque of the knob $\tau$ (low: $3.3\pm 0.6$Nmm, High: $18.8\pm 3.9$Nm). The reaction force is modulated by varying the vertical forces applied on the knob which rests on a plastic surface. 
When the finger is in the soft configuration, the motion is near-identical to the sliding motions introduced in the finger sliding experiment.
In the rigid finger configuration, a secondary motion is programmed to replicate the same motion to ensure the robot doesn't damage itself. 

% Finger gaiting
In the finger gaiting task, a plastic block is held between the thumb and four fingers shown in Fig.\ref{fig:experiment}C.
Starting from all four fingers contacting the block, a finger gaiting pattern is executed (shown by 1, 2, 3, 4) in repeat until eventually the block is dropped. 
Only the width $w$ of the block is varied during this experiment ($\pm 5$mm), while the number of completed gaits and the holding force $F_\mathrm{grip}$ is measured by an inbuilt load cell.

\subsubsection{Grasp type categorization}
\label{sec:meth_grasp_categorization}

The grasp types shown in Fig.\ref{fig:wrist}C for both the robot and human are categorized based on which part of the hand is used to hold/interact with the object and its posture. Each grasp is also related with the grasp taxonomies in\cite{feix2015grasp}.
The \say{Finger surface} grasp (only present for the human) is achieved by keeping the four fingers straight, and using that as a surface to push the object against by the thumb tip (corresponds with \#22:Parallel extension taxonomy).
The \say{Fingertip} grasp uses only the tips of the fingers and thumb to hold the object (corresponds with \#6-8:Prismatic 2-4 finger taxonomy).
The \say{Tip and thumb} grasp uses the fingertips and the middle and/or proximal phalanges (corresponds with \#10:Power disk taxonomy).
The \say{Power (small)} and \say{Power (large)} grasps are both power grasps where one or more phalanges of the fingers/thumb and the palm is used, distinguished based on the diameter of the grasp (corresponds to \#2:Small diameter and \#1:Large diameter taxonomies).

Although only 24 objects are used for the experiment, 25 grasps are recorded because the paper tape generated two distinct grasp types.

% \nolinenumbers

\printbibliography

@inproceedings{xu2016design,
  title={Design of a highly biomimetic anthropomorphic robotic hand towards artificial limb regeneration},
  author={Xu, Zhe and Todorov, Emanuel},
  booktitle={2016 IEEE International Conference on Robotics and Automation (ICRA)},
  pages={3485--3492},
  year={2016},
  organization={IEEE}
}

@article{manti2015bioinspired,
  title={A bioinspired soft robotic gripper for adaptable and effective grasping},
  author={Manti, Mariangela and Hassan, Taimoor and Passetti, Giovanni and D'Elia, Nicol{\`o} and Laschi, Cecilia and Cianchetti, Matteo},
  journal={Soft Robotics},
  volume={2},
  number={3},
  pages={107--116},
  year={2015},
  publisher={Mary Ann Liebert, Inc. 140 Huguenot Street, 3rd Floor New Rochelle, NY 10801 USA}
}

@article{zhou2019soft,
  title={A soft-robotic approach to anthropomorphic robotic hand dexterity},
  author={Zhou, Jianshu and Chen, Xiaojiao and Chang, Ukyoung and Lu, Jui-Ting and Leung, Clarisse Ching Yau and Chen, Yonghua and Hu, Yong and Wang, Zheng},
  journal={IEEE Access},
  volume={7},
  pages={101483--101495},
  year={2019},
  publisher={IEEE}
}

@article{shorthose2022design,
  title={Design of a 3D-printed soft robotic hand with integrated distributed tactile sensing},
  author={Shorthose, Oliver and Albini, Alessandro and He, Liang and Maiolino, Perla},
  journal={IEEE Robotics and Automation Letters},
  volume={7},
  number={2},
  pages={3945--3952},
  year={2022},
  publisher={IEEE}
}

@article{odhner2014compliant,
  title={A compliant, underactuated hand for robust manipulation},
  author={Odhner, Lael U and Jentoft, Leif P and Claffee, Mark R and Corson, Nicholas and Tenzer, Yaroslav and Ma, Raymond R and Buehler, Martin and Kohout, Robert and Howe, Robert D and Dollar, Aaron M},
  journal={The International Journal of Robotics Research},
  volume={33},
  number={5},
  pages={736--752},
  year={2014},
  publisher={SAGE Publications Sage UK: London, England}
}

@article{santello2016hand,
  title={Hand synergies: Integration of robotics and neuroscience for understanding the control of biological and artificial hands},
  author={Santello, Marco and Bianchi, Matteo and Gabiccini, Marco and Ricciardi, Emiliano and Salvietti, Gionata and Prattichizzo, Domenico and Ernst, Marc and Moscatelli, Alessandro and J{\"o}rntell, Henrik and Kappers, Astrid ML and others},
  journal={Physics of life reviews},
  volume={17},
  pages={1--23},
  year={2016},
  publisher={Elsevier}
}

@article{feix2015grasp,
  title={The grasp taxonomy of human grasp types},
  author={Feix, Thomas and Romero, Javier and Schmiedmayer, Heinz-Bodo and Dollar, Aaron M and Kragic, Danica},
  journal={IEEE Transactions on human-machine systems},
  volume={46},
  number={1},
  pages={66--77},
  year={2015},
  publisher={IEEE}
}

@article{kapandji1986clinical,
  title={Clinical test of apposition and counter-apposition of the thumb},
  author={Kapandji, A},
  journal={Annales de chirurgie de la main: organe officiel des societes de chirurgie de la main},
  volume={5},
  number={1},
  pages={67--73},
  year={1986}
}

@article{seminara2023hierarchical,
  title={A hierarchical sensorimotor control framework for human-in-the-loop robotic hands},
  author={Seminara, Lucia and Dosen, Strahinja and Mastrogiovanni, Fulvio and Bianchi, Matteo and Watt, Simon and Beckerle, Philipp and Nanayakkara, Thrishantha and Drewing, Knut and Moscatelli, Alessandro and Klatzky, Roberta L and others},
  journal={Science Robotics},
  volume={8},
  number={78},
  pages={eadd5434},
  year={2023},
  publisher={American Association for the Advancement of Science}
}

@article{kim2019fluid,
  title={Fluid lubricated dexterous finger mechanism for human-like impact absorbing capability},
  author={Kim, Yong-Jae and Yoon, Junsuk and Sim, Young-Woo},
  journal={IEEE Robotics and Automation Letters},
  volume={4},
  number={4},
  pages={3971--3978},
  year={2019},
  publisher={IEEE}
}

@article{gilday2023sensing,
  title={Sensing, actuating, and interacting through passive body dynamics: A framework for soft robotic hand design},
  author={Gilday, Kieran and Hughes, Josie and Iida, Fumiya},
  journal={Soft Robotics},
  volume={10},
  number={1},
  pages={159--173},
  year={2023},
  publisher={Mary Ann Liebert, Inc., publishers 140 Huguenot Street, 3rd Floor New~…}
}

@article{hughes2018anthropomorphic,
  title={An anthropomorphic soft skeleton hand exploiting conditional models for piano playing},
  author={Hughes, JAE and Maiolino, P and Iida, Fumiya},
  journal={Science Robotics},
  volume={3},
  number={25},
  pages={eaau3098},
  year={2018},
  publisher={American Association for the Advancement of Science}
}

@article{gilday2021wrist,
  title={Wrist-driven passive grasping: interaction-based trajectory adaption with a compliant anthropomorphic hand},
  author={Gilday, Kieran and Hughes, Josie and Iida, Fumiya},
  journal={Bioinspiration \& Biomimetics},
  volume={16},
  number={2},
  pages={026024},
  year={2021},
  publisher={IOP Publishing}
}

@article{frost1994perspectives,
  title={Perspectives: a vital biomechanical model of synovial joint design},
  author={Frost, Harold M},
  journal={The Anatomical Record},
  volume={240},
  number={1},
  pages={1--18},
  year={1994},
  publisher={Wiley Online Library}
}

@article{wadsworth1983clinical,
  title={Clinical anatomy and mechanics of the wrist and hand},
  author={Wadsworth, Carolyn T},
  journal={Journal of Orthopaedic \& Sports Physical Therapy},
  volume={4},
  number={4},
  pages={206--216},
  year={1983},
  publisher={JOSPT, Inc. JOSPT, 1033 North Fairfax Street, Suite 304, Alexandria, VA~…}
}

@inproceedings{deimel2013compliant,
  title={A compliant hand based on a novel pneumatic actuator},
  author={Deimel, Raphael and Brock, Oliver},
  booktitle={2013 IEEE International Conference on Robotics and Automation},
  pages={2047--2053},
  year={2013},
  organization={IEEE}
}

@article{deimel2016novel,
  title={A novel type of compliant and underactuated robotic hand for dexterous grasping},
  author={Deimel, Raphael and Brock, Oliver},
  journal={The International Journal of Robotics Research},
  volume={35},
  number={1-3},
  pages={161--185},
  year={2016},
  publisher={SAGE Publications Sage UK: London, England}
}

@article{puhlmann2022rbo,
  title={RBO Hand 3: A platform for soft dexterous manipulation},
  author={Puhlmann, Steffen and Harris, Jason and Brock, Oliver},
  journal={IEEE Transactions on Robotics},
  volume={38},
  number={6},
  pages={3434--3449},
  year={2022},
  publisher={IEEE}
}

@article{bhatt2022surprisingly,
  title={Surprisingly robust in-hand manipulation: An empirical study},
  author={Bhatt, Aditya and Sieler, Adrian and Puhlmann, Steffen and Brock, Oliver},
  journal={arXiv preprint arXiv:2201.11503},
  year={2022}
}

@inproceedings{grioli2012adaptive,
  title={Adaptive synergies: an approach to the design of under-actuated robotic hands},
  author={Grioli, Giorgio and Catalano, Manuel and Silvestro, Emanuele and Tono, Simone and Bicchi, Antonio},
  booktitle={2012 IEEE/RSJ International Conference on Intelligent Robots and Systems},
  pages={1251--1256},
  year={2012},
  organization={IEEE}
}

@article{della2018toward,
  title={Toward dexterous manipulation with augmented adaptive synergies: The pisa/iit softhand 2},
  author={Della Santina, Cosimo and Piazza, Cristina and Grioli, Giorgio and Catalano, Manuel G and Bicchi, Antonio},
  journal={IEEE Transactions on Robotics},
  volume={34},
  number={5},
  pages={1141--1156},
  year={2018},
  publisher={IEEE}
}

@article{pfeifer2006designing,
  title={Designing Intelligent Robots—On the Implications of Embodiment—},
  author={Pfeifer, Rolf and Iida, Fumiya and Gomez, Gabriel},
  journal={Journal of the Robotics Society of Japan},
  volume={24},
  number={7},
  pages={783--790},
  year={2006},
  publisher={The Robotics Society of Japan}
}

@article{pfeifer2007self,
  title={Self-organization, embodiment, and biologically inspired robotics},
  author={Pfeifer, Rolf and Lungarella, Max and Iida, Fumiya},
  journal={science},
  volume={318},
  number={5853},
  pages={1088--1093},
  year={2007},
  publisher={American Association for the Advancement of Science}
}

@article{vazhapilli2019systematic,
  title={A systematic approach to evaluating and benchmarking robotic hands—the ffp index},
  author={Vazhapilli Sureshbabu, Anand and Metta, Giorgio and Parmiggiani, Alberto},
  journal={Robotics},
  volume={8},
  number={1},
  pages={7},
  year={2019},
  publisher={MDPI}
}

@article{dasari2021rb2,
  title={RB2: Robotic Manipulation Benchmarking with a Twist},
  author={Dasari, Sudeep and Wang, Jianren and Hong, Joyce and Bahl, Shikhar and Lin, Yixin and Wang, Austin S and Thankaraj, Abitha and Singh Chahal, Karanbir and Calli, Berk and Gupta, Saurabh and others},
  journal={NeurIPS 2021 Datasets and Benchmarks Track},
  year={2021}
}

@article{mnyusiwalla2020bin,
  title={A bin-picking benchmark for systematic evaluation of robotic pick-and-place systems},
  author={Mnyusiwalla, Hussein and Triantafyllou, Pavlos and Sotiropoulos, Panagiotis and Roa, M{\'a}ximo A and Friedl, Werner and Sundaram, Ashok M and Russell, Duncan and Deacon, Graham},
  journal={IEEE Robotics and Automation Letters},
  volume={5},
  number={2},
  pages={1389--1396},
  year={2020},
  publisher={IEEE}
}

@article{cruciani2020benchmarking,
  title={Benchmarking in-hand manipulation},
  author={Cruciani, Silvia and Sundaralingam, Balakumar and Hang, Kaiyu and Kumar, Vikash and Hermans, Tucker and Kragic, Danica},
  journal={IEEE Robotics and Automation Letters},
  volume={5},
  number={2},
  pages={588--595},
  year={2020},
  publisher={IEEE}
}

@article{bourquin2004self,
  title={Self-organization of locomotion in modular robots},
  author={Bourquin, Yvan and Ijspeert, Auke Jan and Harvey, Inman},
  journal={Unpublished Diploma Thesis, http://birg. epfl. ch/page53073. html},
  year={2004}
}

@article{stella2023paws,
  title={PAWS: A Synergy-Based Robotic Quadruped Leveraging Passivity for Robustness and Behavioural Diversity},
  author={Stella, Francesco and Achkar, Mickael and Della Santina, Cosimo and Hughes, Josie},
  year={2023}
}

@article{badri2022birdbot,
  title={BirdBot achieves energy-efficient gait with minimal control using avian-inspired leg clutching},
  author={Badri-Spr{\"o}witz, Alexander and Aghamaleki Sarvestani, Alborz and Sitti, Metin and Daley, Monica A},
  journal={Science Robotics},
  volume={7},
  number={64},
  pages={eabg4055},
  year={2022},
  publisher={American Association for the Advancement of Science}
}

@inproceedings{calli2015ycb,
  title={The ycb object and model set: Towards common benchmarks for manipulation research},
  author={Calli, Berk and Singh, Arjun and Walsman, Aaron and Srinivasa, Siddhartha and Abbeel, Pieter and Dollar, Aaron M},
  booktitle={2015 international conference on advanced robotics (ICAR)},
  pages={510--517},
  year={2015},
  organization={IEEE}
}

@article{rosenbaum2012cognition,
  title={Cognition, action, and object manipulation.},
  author={Rosenbaum, David A and Chapman, Kate M and Weigelt, Matthias and Weiss, Daniel J and van der Wel, Robrecht},
  journal={Psychological bulletin},
  volume={138},
  number={5},
  pages={924},
  year={2012},
  publisher={American Psychological Association}
}

@article{murphy2018structure,
  title={Structure, function, and control of the human musculoskeletal network},
  author={Murphy, Andrew C and Muldoon, Sarah F and Baker, David and Lastowka, Adam and Bennett, Brittany and Yang, Muzhi and Bassett, Danielle S},
  journal={PLoS biology},
  volume={16},
  number={1},
  pages={e2002811},
  year={2018},
  publisher={Public Library of Science San Francisco, CA USA}
}

@article{feix2014analysis,
  title={Analysis of human grasping behavior: Object characteristics and grasp type},
  author={Feix, Thomas and Bullock, Ian M and Dollar, Aaron M},
  journal={IEEE transactions on haptics},
  volume={7},
  number={3},
  pages={311--323},
  year={2014},
  publisher={IEEE}
}

@inproceedings{weghe2004act,
  title={The ACT hand: Design of the skeletal structure},
  author={Weghe, M Vande and Rogers, Matthew and Weissert, Michael and Matsuoka, Yoky},
  booktitle={IEEE International Conference on Robotics and Automation, 2004. Proceedings. ICRA'04. 2004},
  volume={4},
  pages={3375--3379},
  year={2004},
  organization={IEEE}
}

@inproceedings{toshimitsu2023getting,
  title={Getting the Ball Rolling: Learning a Dexterous Policy for a Biomimetic Tendon-Driven Hand with Rolling Contact Joints},
  author={Toshimitsu, Yasunori and Forrai, Benedek and Cangan, Barnabas Gavin and Steger, Ulrich and Knecht, Manuel and Weirich, Stefan and Katzschmann, Robert K},
  booktitle={2023 IEEE-RAS 22nd International Conference on Humanoid Robots (Humanoids)},
  pages={1--7},
  year={2023},
  organization={IEEE}
}

@article{dollar2006robust,
  title={A robust compliant grasper via shape deposition manufacturing},
  author={Dollar, Aaron M and Howe, Robert D},
  journal={IEEE/ASME transactions on mechatronics},
  volume={11},
  number={2},
  pages={154--161},
  year={2006},
  publisher={IEEE}
}

@article{miriyev2020skills,
  title={Skills for physical artificial intelligence},
  author={Miriyev, Aslan and Kova{\v{c}}, Mirko},
  journal={Nature Machine Intelligence},
  volume={2},
  number={11},
  pages={658--660},
  year={2020},
  publisher={Nature Publishing Group UK London}
}

@inproceedings{bern2022simulation,
  title={Simulation and fabrication of soft robots with embedded skeletons},
  author={Bern, James M and Zargarbashi, Fatemeh and Zhang, Annan and Hughes, Josie and Rus, Daniela},
  booktitle={2022 International Conference on Robotics and Automation (ICRA)},
  pages={5205--5211},
  year={2022},
  organization={IEEE}
}

@article{cculha2016enhancement,
  title={Enhancement of finger motion range with compliant anthropomorphic joint design},
  author={{\c{C}}ulha, Utku and Iida, Fumiya},
  journal={Bioinspiration \& biomimetics},
  volume={11},
  number={2},
  pages={026001},
  year={2016},
  publisher={IOP Publishing}
}

@article{nanayakkara2017role,
  title={The role of morphology of the thumb in anthropomorphic grasping: a review},
  author={Nanayakkara, Visakha K and Cotugno, Giuseppe and Vitzilaios, Nikolaos and Venetsanos, Demetrios and Nanayakkara, Thrishantha and Sahinkaya, M Necip},
  journal={Frontiers in mechanical engineering},
  volume={3},
  pages={5},
  year={2017},
  publisher={Frontiers Media SA}
}

@article{ijspeert2007swimming,
  title={From swimming to walking with a salamander robot driven by a spinal cord model},
  author={Ijspeert, Auke Jan and Crespi, Alessandro and Ryczko, Dimitri and Cabelguen, Jean-Marie},
  journal={science},
  volume={315},
  number={5817},
  pages={1416--1420},
  year={2007},
  publisher={American Association for the Advancement of Science}
}

@article{khadivar2022emg,
  title={EMG-driven shared human-robot compliant control for in-hand object manipulation in hand prostheses},
  author={Khadivar, Farshad and Mendez, Vincent and Correia, Carolina and Batzianoulis, Iason and Billard, Aude and Micera, Silvestro},
  journal={Journal of Neural Engineering},
  volume={19},
  number={6},
  pages={066024},
  year={2022},
  publisher={IOP Publishing}
}

@inproceedings{gilday2022intelligent,
  title={Intelligent Soft Hands and Benchmarking towards General-Purpose Robotic Manipulation},
  author={Gilday, Kieran and Iida, Fumiya},
  booktitle={IOP Conference Series: Materials Science and Engineering},
  volume={1261},
  number={1},
  pages={012010},
  year={2022},
  organization={IOP Publishing}
}

@inproceedings{franzese2021ilosa,
  title={ILoSA: Interactive learning of stiffness and attractors},
  author={Franzese, Giovanni and M{\'e}sz{\'a}ros, Anna and Peternel, Luka and Kober, Jens},
  booktitle={2021 IEEE/RSJ International Conference on Intelligent Robots and Systems (IROS)},
  pages={7778--7785},
  year={2021},
  organization={IEEE}
}

@article{piazza2019century,
  title={A century of robotic hands},
  author={Piazza, C and Grioli, G and Catalano, MG and Bicchi, AJAROC},
  journal={Annual Review of Control, Robotics, and Autonomous Systems},
  volume={2},
  pages={1--32},
  year={2019},
  publisher={Annual Reviews}
}

% \linenumbers

\clearpage

\section*{Author contributions}
K.J and J.H conceived the idea and designed the research. K.J developed the robotic systems, performed the experiment, and analyzed the data. K.J and J.H wrote the manuscript. 

\section*{Competing interests}
The authors declare no competing interests in this work.

\nolinenumbers
\clearpage

\setcounter{page}{1}

\section*{\large Supplementary Material} 
\renewcommand\thesection{S\arabic{section}}
\setcounter{section}{0}
\section{Links to external resources}
\label{suppsec:resource}
The uncut video of the extended pick and place operation (section \ref{sec:extended_opeartion}) can be found \href{https://www.youtube.com/watch?v=jIvlIkWIGKw&ab_channel=CreateLab}{here}.
All the code used to operate the ADAPT Hand can be found \href{https://gitlab.epfl.ch/create-lab/bio-inspired-robot-hand/adapt-hand}{here}. 
The CAD model of the ADAPT Hand can be found \href{https://a360.co/3TkZH4T}{here}. 
All data collected from the experiments alongside any other questions should be forwarded to the corresponding author.

\end{document}